\documentclass{midl} 

\usepackage{mwe} 
\usepackage{booktabs}
\usepackage{rotating}
\usepackage{array}
\usepackage{bm}
\usepackage{todonotes}
\newcommand{\bb}[1]{\bm{\mathrm{#1}}}
\newcolumntype{L}[1]{>{\raggedright\arraybackslash}p{#1}}
\usepackage{tikz}
\newcolumntype{Y}{>{\raggedright\arraybackslash}X} 

\jmlryear{2019}
\jmlrworkshop{Full Paper -- MIDL 2019}

\title[Learning beamforming in ultrasound imaging]{Learning beamforming in ultrasound imaging}

\midlauthor{\Name{Sanketh Vedula\nametag{$^{1}$}\midljointauthortext{Contributed equally}} \Email{sanketh@cs.technion.ac.il}\\
\Name{Ortal Senouf\nametag{$^{1}$}\midlotherjointauthor} \Email{senouf@campus.technion.ac.il}\\
\Name{Grigoriy Zurakhov\nametag{$^{1}$}}
\Email{grishaz@campus.technion.ac.il}\\
\Name{Alex Bronstein\nametag{$^{1}$}} \Email{bron@cs.technion.ac.il}\\
\Name{Oleg Michailovich\nametag{$^{2}$}} \Email{olegm@uwaterloo.ca}\\
\Name{Michael Zibulevsky\nametag{$^{1}$}} \Email{mzib@cs.technion.ac.il}\\
\addr $^{1}$ Technion, Israel\\
\addr $^{2}$ University of Waterloo, Canada
}

\begin{document}

\maketitle

\begin{abstract}
Medical ultrasound (US) is a widespread imaging modality owing its popularity to cost efficiency, portability, speed, and lack of harmful ionizing radiation. In this paper, we demonstrate that replacing the traditional ultrasound processing pipeline with a data-driven, learnable counterpart leads to significant improvement in image quality. Moreover, we demonstrate that greater improvement can be achieved through a learning-based design of the transmitted beam patterns simultaneously with learning an image reconstruction pipeline. We evaluate our method on an in-vivo first-harmonic cardiac ultrasound dataset acquired from volunteers and demonstrate the significance of the learned pipeline and transmit beam patterns on the image quality when compared to standard transmit and receive beamformers used in high frame-rate US imaging. We believe that the presented methodology provides a fundamentally different perspective on the classical problem of ultrasound beam pattern design.
\end{abstract}


\section{Introduction}

Recently, there has been a surge of interest in applying learning-based techniques to improve ultrasound imaging. In \cite{senouf2018high} and \cite{vedula2018high}, we demonstrated that convolutional neural networks (CNNs) can be employed to reconstruct high-quality images acquired through high-framerate ultrasound acquisition protocols. Similarly, in \cite{gasse2017plane}, the authors proposed that CNNs could be used as a means to perform plane-wave compounding requiring significantly lesser number of plane-waves to reconstruct a high-quality image. \cite{simson2018Deepformer} proposed to approximate time-consuming beamformers such as minimum-variance beamforming using CNNs. In \cite{DNNBF2018}, the authors proposed to use process time-delayed and phase-rotated signals using fully connected networks showing to improve ultrasound image reconstruction. 
Apart from ultrasound image formation, CNNs were used in ultrasound post-processing for real-time despeckling and CT-quality image reconstruction \cite{vedula2017towards}, for speed-of-sound estimation \cite{Feigin2018DLSoS} and for ultrasound segmentation directly from the raw-data \cite{nair2018deep}.

\paragraph{Contributions.} Viewing US imaging as an inverse problem, in which a latent image is reconstructed from a set of measurements, the above mentioned studies focused on learning (parts of) the inverse operator producing an image from the measurements. The scope of the present paper differs sharply in the sense that we propose to learn the parameters of the \emph{forward model}, specifically, the transmitted patterns. We propose to jointly learn the end-to-end transmit (Tx) and receive (Rx) beamformers optimized for the task of high-framerate ultrasound imaging, in which the number of measurements per image has a direct impact on the frame rate. We demonstrate a significant improvement in the image quality compared to the standard patterns used in this setting.

Unlike our previous works \cite{senouf2018high, vedula2018high} that train separate networks for the \textit{in-phase} (I) and \textit{quadrature} (Q) components of the demodulated received ultrasound data, we propose a unified dual-pathway network that trains jointly I and Q minimizing for the loss defined on the final envelope image (Figure \ref{fig:USpipelineArch}).   
We also propose a new beamforming layer inspired by \cite{jaderberg2015spatial}, that implements beamforming as a \textit{differentiable} geometric transformation between pre-beamformed Rx signal and the beamformed one. This results in a fully-differentiable end-to-end Rx beamforming and signal processing pipeline that can be easily generalized to a variety of imaging settings.
By rendering the end-to-end Rx pipeline differentiable, we demonstrate that the Tx protocols can be optimized together with the Rx beamforming and reconstruction pipeline, leading to significant improvement in image quality. To the best of our knowledge, this is the first time simultaneous end-to-end learning of hardware parameters and signal processing algorithms are used in US imaging.

\section{Methods}

\begin{figure}
    \centering
    \includegraphics[width=0.75\textwidth]{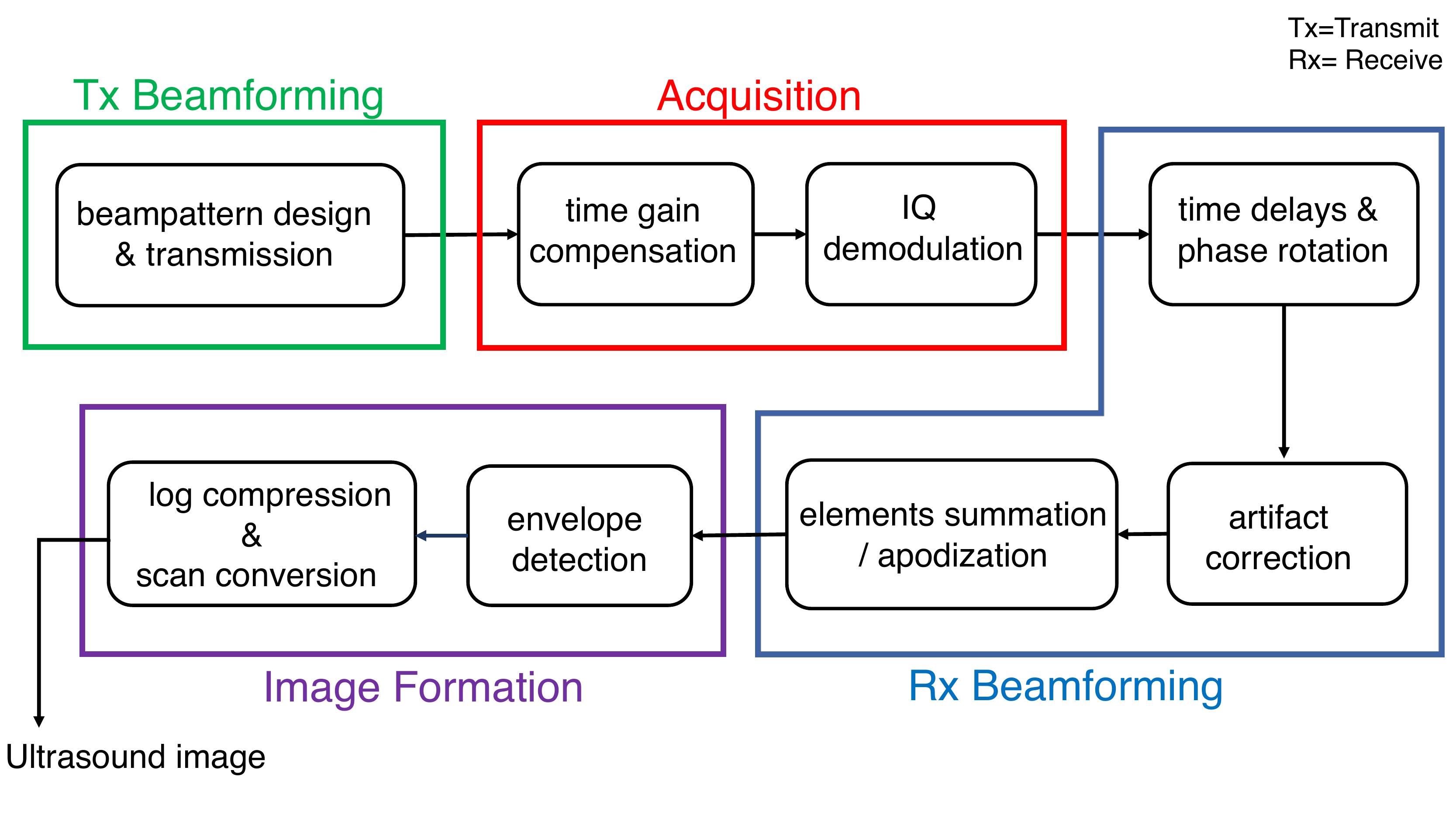} 
    \caption{The ultrasound imaging pipeline}
    \label{fig:USpipelineArch}
\end{figure}

Traditionally, a US imaging pipeline consists of the following stages: Tx beamforming, acquisition, Rx beamforming, and image formation. In Tx beamforming, depending on the desired frame-rate and quality, a suitable number of transmissions and their corresponding beam profile are chosen and the piezo-electric transducers are programmed accordingly to transmit the beams. Post-transmission, the echoes are received by the same transducer array; these signals are demodulated and focused by applying the appropriate time-delays and phase-rotations to produce the beamformed signal. The beamformed signal is further processed to correct the artifacts (if acquired through high frame-rate transmit modes) and apodized to suppress the side-lobes. We refer to these stages of processing the demodulated signals collectively as \textit{Rx beamforming} (Figure \ref{fig:USpipelineArch}). After Rx beamforming, the envelope is extracted from the complex signal, followed by a log-compression and scan-conversion to produce the final ultrasound image.

\subsection{Learned end-to-end Rx pipeline}

\begin{figure}
    \centering
    \includegraphics[width=0.98\textwidth]{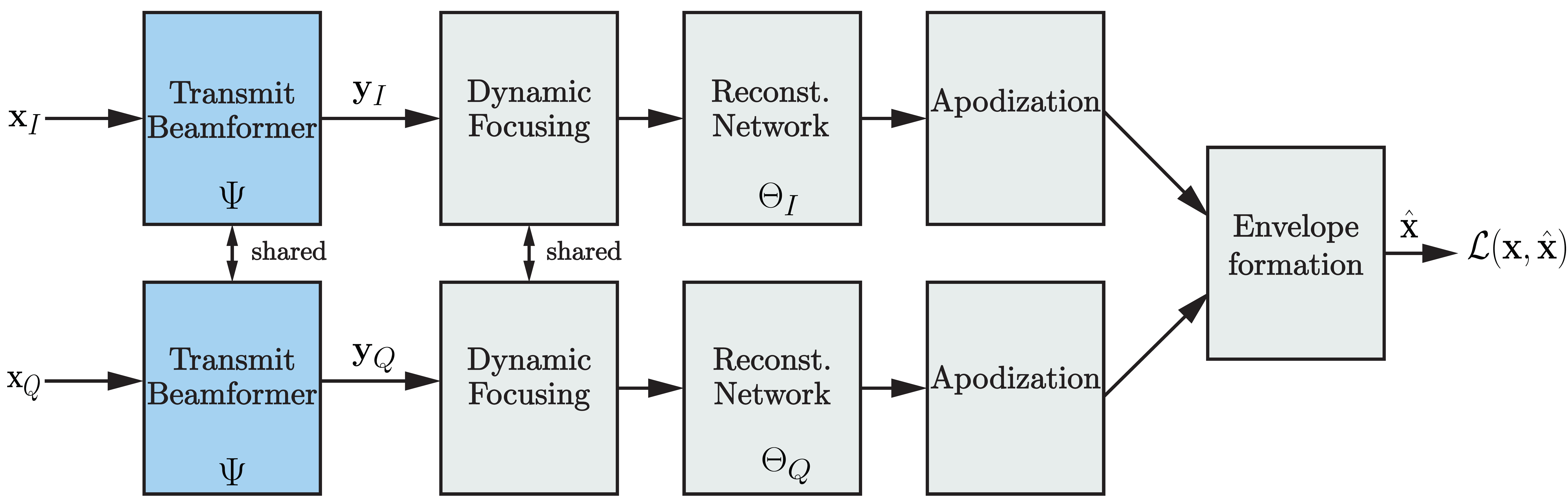}     
    \caption{Learned end-tco-end Tx-Rx pipeline. The stages: dyanamic focusing, reconstruction network, apodization and envelope formation are together referred to as Rx beamforming.}
    \label{fig:USpipelineArch_proposed}
\end{figure}

In our previous studies \cite{senouf2018high,vedula2018high}, we have used a symmetric encoder-decoder multi-resolution neural network in order to fix the distorted received US signal and get the higher quality undistorted signal. Two networks were trained separately for the I and Q signals, mostly due to computational and technical difficulties to train one network for both. In this paper, we present an architecture that comprises two separate paths for I and Q followed by a layer forming the envelope signal, on which the loss is calculated. $\Theta_I$ and $\Theta_Q$ in Figure \ref{fig:USpipelineArch_proposed} denote the parameters of the two encoder-decoder networks with an architecture similar to that of a U-Net \cite{ronneberger2015u}. Moreover, in our previous works we have trained and applied the networks to the time-delayed and phase rotated signals, which would not allow us to perform manipulations on transmission (Tx) patterns. In this work, we have implemented a time-delays and phase rotation stage (referred to as \textit{dynamic focusing}) in the network architecture, which allows to work on the pre-Rx-beamformed signals directly, as described in Figure \ref{fig:USpipelineArch_proposed}. 

Performing time-delays and phase-rotations through convolutions is not trivial because it would require a very large support of surrounding data points. This, in turn would require a computationally intractable number of arithmetic operations to approximate the delays. In order to overcome this problem, we propose to perform time-delays and phase-rotations as a differentiable geometric transformation of the pre-beamformed signal. We introduce a spatial transformation layer inspired by the works of \cite{jaderberg2015spatial} and \cite{skafte2018deep}, in which the authors proposed a differentiable sampling and interpolation method in order to train and apply affine and, more generally, diffeomorphic transformations to the input. Here, we apply the explicit time delays and phase-rotation (\textit{dynamic focusing}) in a similar fashion. Given the raw signal ${\phi}_{m}(t,\alpha)$ corresponding to focused beams direction $\alpha$ read out from the $m$-th array element at location $\delta_m$ and time $t$, we construct the time-delayed signal as $\hat{\phi}_m(t,\alpha) =\phi_m(\hat{t}, \alpha)$,
where
$$
\label{eq:time_delays}
\hat{t} = \frac{t}{2}+\sqrt[]{\frac{t^2}{4}-t\sin \alpha \frac{\delta_m}{c}+\bigg(\frac{\delta_m}{c}\bigg)^2},
$$ 
and $c$ is the speed of sound in the tissue, assumed to be $1540$ m/s. In addition, in order to eliminate phase error, phase rotation is applied to the complex signal in its explicit form, as described in \cite{chang1993phase}:
$$
\label{eq:phase rotation}
\left( \begin{array}{c}
\Re\, \mathrm{IQ} \\ 
\Im\, \mathrm{IQ}
\end{array}
\right) 
=
\left( \begin{array}{cc}
\cos(\omega_0 (\hat{t}-t) ) & -\sin(\omega_0 (\hat{t}-t) ) \\
\sin(\omega_0 (\hat{t}-t) ) & \cos(\omega_0 (\hat{t}-t) )  
\end{array}
\right) 
\left( \begin{array}{c}
\Re\, \hat{\phi}_m(t,\alpha) \\ 
\Im\, \hat{\phi}_m(t,\alpha)
\end{array}
\right), 
$$
where $\omega_0$ is the modulation frequency and $\Re$ and $\Im$ denote, respectively, the real and imaginary parts of a complex number.

The \textit{dynamic focusing} is placed after the \textit{Tx beamformer} layer and before the reconstruction network, as depicted in Figure \ref{fig:USpipelineArch_proposed}. While in our implementation, all the parameters defining the time delay and phase rotation transformations are fixed, they can be trained as well.

\subsection{Learning optimal transmit patterns}

The problem of learning optimal transmitted patterns together with Rx beamforming and reconstruction can be formulated as a simultaneous learning of the forward model and its (approximate) inverse.  
Ultrasound imaging can be viewed end-to-end as a process that given a latent image $\bb{x}$ (the object being imaged) generates a set of measurements $\bb{y}$ thereof by sampling from a parametric conditional distribution $\bb{y} \sim p_\psi (\bb{y} | \bb{x})$. This conditional distribution is known as the likelihood in the Bayesian jargon, and can be viewed as a stochastic forward model. The set of parameters $\psi$ denotes collectively the settings of the imaging hardware, including the patterns transmitted to obtain the measurements.

The goal of the signal processing pipeline is to produce the an estimate $\hat{\bb{x}}$ of the latent image $\bb{x}$ given the measurements $\bb{y}$. We denote the estimator as $\hat{\bb{x}}_\theta (y)$ and refer to it as the inverse operator, implying that it should invert the action of the forward model. The set of parameters $\theta$ denotes the trainable degrees of freedom of the reconstruction pipeline; in our case, these are the weights of the reconstruction neural network. 
We propose to simultaneously learn the parameters of both the forward model and the inverse operator such as to optimize performance in a specific task.
This can be carried out by minimizing the expected loss,
$$
\min_{\mathbf{\theta},\mathbf{\psi}} \, \mathbb{E}_{\mathbf{x}\sim p(\mathbf{x})}  \, \mathbb{E}_{\mathbf{y} \sim p_{\mathbf{\psi}}(\mathbf{y} | \mathbf{x}) } \,\, \mathcal{L}(\hat{\mathbf{x}}_{\mathbf{\theta}}( \mathbf{y} ), \mathbf{x}),
$$
where $\mathcal{L}(\hat{\mathbf{x}},\mathbf{x})$ measures the discrepancy between the ground truth image $\mathbf{x}$ and its estimate $\hat{\mathbf{x}}$. In practice, the expectations are replaced by finite-sample approximation on the training set. Note that the expectation taken over $\mathbf{y} \sim p_{\mathbf{\psi}}(\mathbf{y} | \mathbf{x})$ embodies the parametric forward model whose parameters $\mathbf{\psi}$ (reflecting the transmission pattern) are optimized simultaneously with the parameters of the inverse operator (i.e., the computational process applied to the measurement $\mathbf{y}$ to recover the latent signal), in our case, the reconstruction network. This training regime resembles in spirit the training of autoencoder networks; in our case, the architecture of the encoder is fixed as dictated by the imaging hardware, and only parameters under the user's control can be trained.

The idea of simultaneously training a signal reconstruction process and some parameters of the signal acquisition forward model has been previously corroborated in computational imaging, including compressed tomography~\cite{Menashe2014}, phase-coded aperture extended depth-of-field and range image sensing~\cite{Haim2018IEEETCI}. In all the mentioned cases, a significant improvement in performance was observed both in simulation and in real systems. 

In our current work, we refer only to first harmonic ultrasound imaging, whose forward model is linear. This means that applying manipulations to the received signal is equivalent to applying them on the transmitted signal, as has been shown in \cite{prieur2013correspondence}. This way the forward model is parameterized by a set of linear combinations of the original received beam,
$$
\mathbf{y}_j={\sum_{i=1}^{L} \psi_{ij} \mathbf{x}_i}, \quad \{\mathbf{y}_j\}_{j=1}^{M}
$$
where $L$ is the number of the original received beams, $M$ is the number of new learned beams, and the matrix $\psi$ encodes the transmit beam patterns. It has been shown \cite{prieur2013correspondence} that this approach can faithfully emulate measurements that would be formed from a more complex excitation.

\section{Experiments and discussion}

\subsection{Data acquisition}

The FOV was scanned by $140/140$ Tx/Rx lines, each of them covered a sector of $0.54^\circ$. We refer to this baseline acquisition scenario as \textit{single-line acquisition} (SLA) and consider it to be the ground truth in all reduced transmission experiments.
In order to assess the generalization performance of our method, we used a cine loop from a patient whose data were excluded from the training/validation set. 

\subsection{Settings}
In order to evaluate the contribution of the joint training of the transmit pattern and the received signal reconstruction, we have designed a two-stage experiment. First, we trained only the reconstruction network and fixed the Tx beamforming parameters. Second, we used a pre-convergence checkpoint of the reconstruction network as a starting point for the joint training. At this stage, we also trained the Tx parameters. In order to factor out the influence of the optimization algorithm, we trained the reconstruction network in both stages with the same optimizer (Adam, initial learning rate $=0.005$). The Tx parameters were trained using the momentum optimizer with a decaying learning rate  (initial learning rate $=0.005$). The loss function, $\mathcal{L}(\hat{\mathbf{x}},\mathbf{x})$, was set to the $L_1$ error. 

\paragraph{Different initializations.}
We performed the two-stage experiment with different initializations for the Tx parameters using known reduced transmission methods as well as random initialization. We fixed the decimation factor to $10$, meaning that instead of the $140$ original acquisitions, only $14$ measurements were emulated and provided to the reconstruction network. One initialization method was the \textit{multi-line acquisition} (MLA) in which for every wide transmitted beam, $10$ (as the decimation factor) Rx narrow beams are reconstructed. Each $10-$MLA acquisition is emulated by averaging over $10$ consecutive \textit{single-line acquisition} (SLA) Rx signals (as depicted in Figure \ref{fig:MLAMLT} in the Appendix) \cite{rabinovich2013multi}. Another initialization method is the \textit{multi-line transmission} (MLT) in which a comb of uniformly spaced narrow beams is transmitted simultaneously. Each $10-$MLT acquisition is emulated by summing over $10$ uniformly spaced received Rx signals from SLA (as presented in Figure \ref{fig:MLAMLT}, in the Appendix) \cite{rabinovich2015multi}. Finally, a random initialization was used to emulate, in a way, a plane wave excitation \cite{montaldo2009coherent}, in which there is no directivity to the beam pattern. In this experiment, mentioned in this paper as $10-$random, $14$ acquisitions of distinct random patterns were emulated.

\paragraph{Different decimation rates.}
In this experiment, we fixed the initialization to MLA and performed the above described two-stage experiment over different decimation rates $7$, $10$, and $20$.

\subsection{Results and discussion}
\paragraph{Notation.} For all the experiments presented within the paper, \textit{Learned Rx} refers to the setting where the transmission is fixed and the reconstruction network alone is trained and \textit{Learned Tx-Rx} refers to the setting in which the transmission patterns are jointly learned with with the reconstruction network. \textit{Fixed Tx -- DAS} refers to the setting where the fixed transmissions are beamformed using a standard delay-and-sum (DAS) beamformer, and \textit{Learned Tx -- DAS} is the setting where learned transmissions are beamformed using a delay-and-sum Rx beamformer.

\begin{figure}
    \centering
    \includegraphics[width=0.32\textwidth]{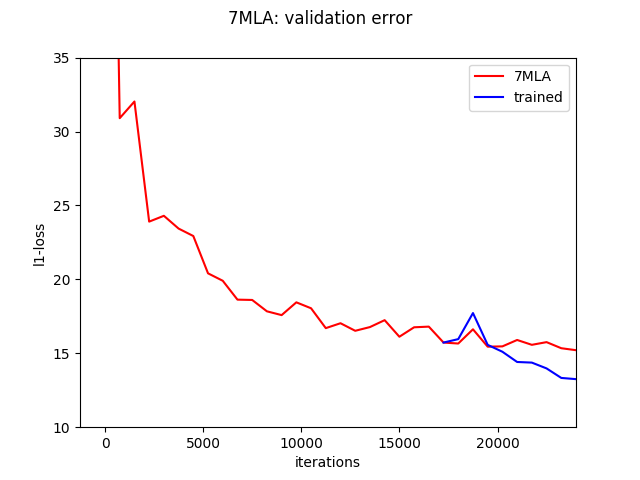}
    \includegraphics[width=0.32\textwidth]{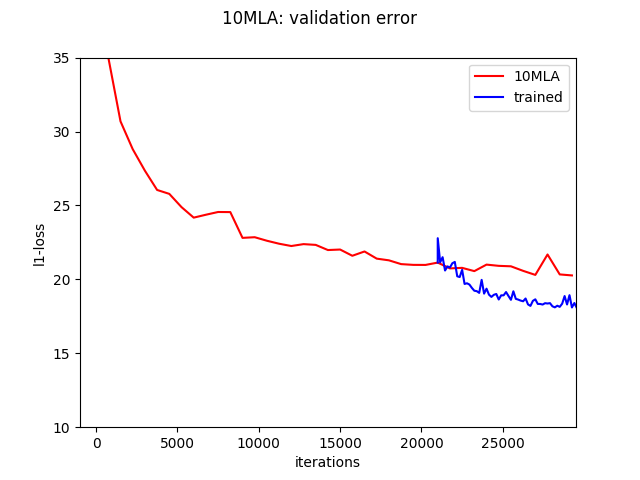}
    \includegraphics[width=0.32\textwidth]{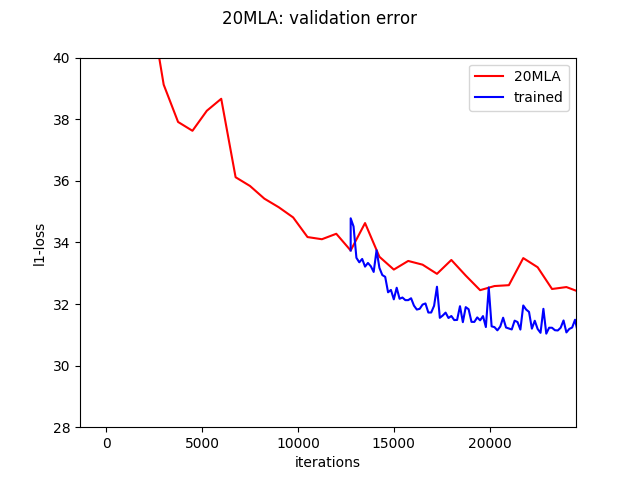}
    \caption{\small Convergence plots. Depicted from left to right are the validation error plots of $7-$, $10-$ and $20-$MLA. The red and blue lines indicate the \emph{learning Rx} and \emph{learning Tx-Rx} settings, respectively.}
    \label{fig:convergence_plots}
\end{figure}

\paragraph{Convergence.} Figure \ref{fig:convergence_plots} displays the validation error plot of the two stages training for the different decimation rates experiment. Each iteration corresponds with a mini-batch, which in our settings its size has been set to one. The error gap between the the learned-Rx and the jointly learned Tx-Rx, in favour of the latter, supports our claim for the superiority of joint learning of forward and inverse models in the case of US acquisition. A similar behaviour was observed for other initializations.    

\paragraph{Train + test split.} We generated a dataset for training the network using cardiac data from six patients; each patient contributed $4-5$ cine loops containing $32$ frames each. The networks were trained on the cineloops of five patients and the testset consists of the cineloops from the patient that was excluded from the trainset. The total trainset consisted of $745$ frames, while the testset consisted of $160$ frames. 

\paragraph{Quantitative results.}  We present the quantitative evaluation of the first cineloop ($32$ frames) in Table \ref{tab:quantitaive_analysis}, the quantitative results for the rest of the cineloops are summarized in the supplementary material\footnote{Supplementary material: available \href{https://vista.cs.technion.ac.il/wp-content/uploads/2019/04/suppmat_VedSenZurBroZibMicMIDL2019.pdf}{here}.}. Table \ref{tab:quantitaive_analysis} (top), summarizing the average quality measures for the different decimation rates, shows improved performance in the sense of the $L_1$ error used to train the models, and in the sense of the peak signal-to-noise ratio (PSNR), which is correlated to the $L_1$ loss. It is interesting to observe that an improvement was also observed in the sense of the structure-similarity (SSIM) measure, for which the models were not trained. In Tables $1$ and $2$, we can observe that the learned Rx pipeline performs significantly better than the fixed Tx with a DAS beamformer. Similar behavior can be observed in all the experiments. More interestingly, one can see that the learned transmissions perform better than the fixed ones even with the DAS beamformer. The best performance, with a significant margin, is achieved when the transmit patterns and the Rx beamformer are jointly learned, in all settings. Comparison between different initializations of transmission patterns for a fixed decimation factor is presented in Table \ref{tab:quantitaive_analysis} (bottom). Observe that the transmission pattern initialized with MLA performs better than MLT and random initializations, also by a significant margin. 

\begin{table}

\centering
\setlength\tabcolsep{1.5pt} 
\begin{tabular}{l||lll|lll|lll}

              &      & 7-MLA &          &      & 10-MLA &          &      & 20-MLA &           \\ 
\hline
              & PSNR & SSIM  & L1-error & PSNR & SSIM   & L1-error & PSNR & SSIM   & L1-error  \\
Fixed Tx -- DAS    & 33.76    & 0.955     & --       & 32.34    & 0.941      & --        & 29.6    & 0.91      & --         \\
Learned Tx -- DAS    & 34.03    & 0.96     & --        & 32.73    & 0.95      & --        & 29.87    & 0.916      & --         \\
Learned Rx    & 42.56    & 0.987     & 19.14        & 39.56    & 0.975      & 24.31        & 35.02    & 0.924      & 38.36         \\
Learned Tx-Rx & 43.4    & 0.99     & 15.94        & 39.98    & 0.98      & 22.19        & 35.32    & 0.95      & 36.24 
\vspace{0.5cm}
\end{tabular}

\begin{tabular}{l||lll|lll|lll}
              &      & 10-MLA &          &      & 10-MLT &          &      & 10-random &           \\ 
\hline
              & PSNR & SSIM  & L1-error & PSNR & SSIM   & L1-error & PSNR & SSIM   & L1-error  \\
Fixed Tx -- DAS    & 32.34    & 0.941      & --         & 24.39    & 0.855      & --       & 24.26    & 0.865      & --         \\
Learned Tx -- DAS    & 32.73    & 0.95      & --       & 25.22   & 0.878      & --        & 25.34    & 0.88      & --         \\              
Learned Rx    & 39.56  & 0.975 & 24.31 & 33.66   & 0.92      & 47.99       & 34.7 & 0.935  & 46.7        \\
Learned Tx-Rx & 39.58    & 0.98     & 22.19 & 35.04    & 0.92      & 41        & 36.52 & 0.95  & 38       
\end{tabular}
\caption{Comparison of average PSNR, SSIM and $L_1$ error measures between different decimation rates of transmissions (top) and different initializations (bottom). First and second rows indicate the performance of fixed and learned transmissions with a standard delay-and-sum (DAS) beamformer, respectively. Third and fourth rows indicate the results corresponding to learned Rx and learned Tx-Rx experiment settings, respectively.}
\label{tab:quantitaive_analysis}
\end{table}

\begin{figure}[ht]
\begin{minipage}[]{\linewidth}
	\begin{tabular}{ c@{\hskip 0.001\textwidth}c@{\hskip 0.001\textwidth}c} 
		\includegraphics[width = 0.33\textwidth]{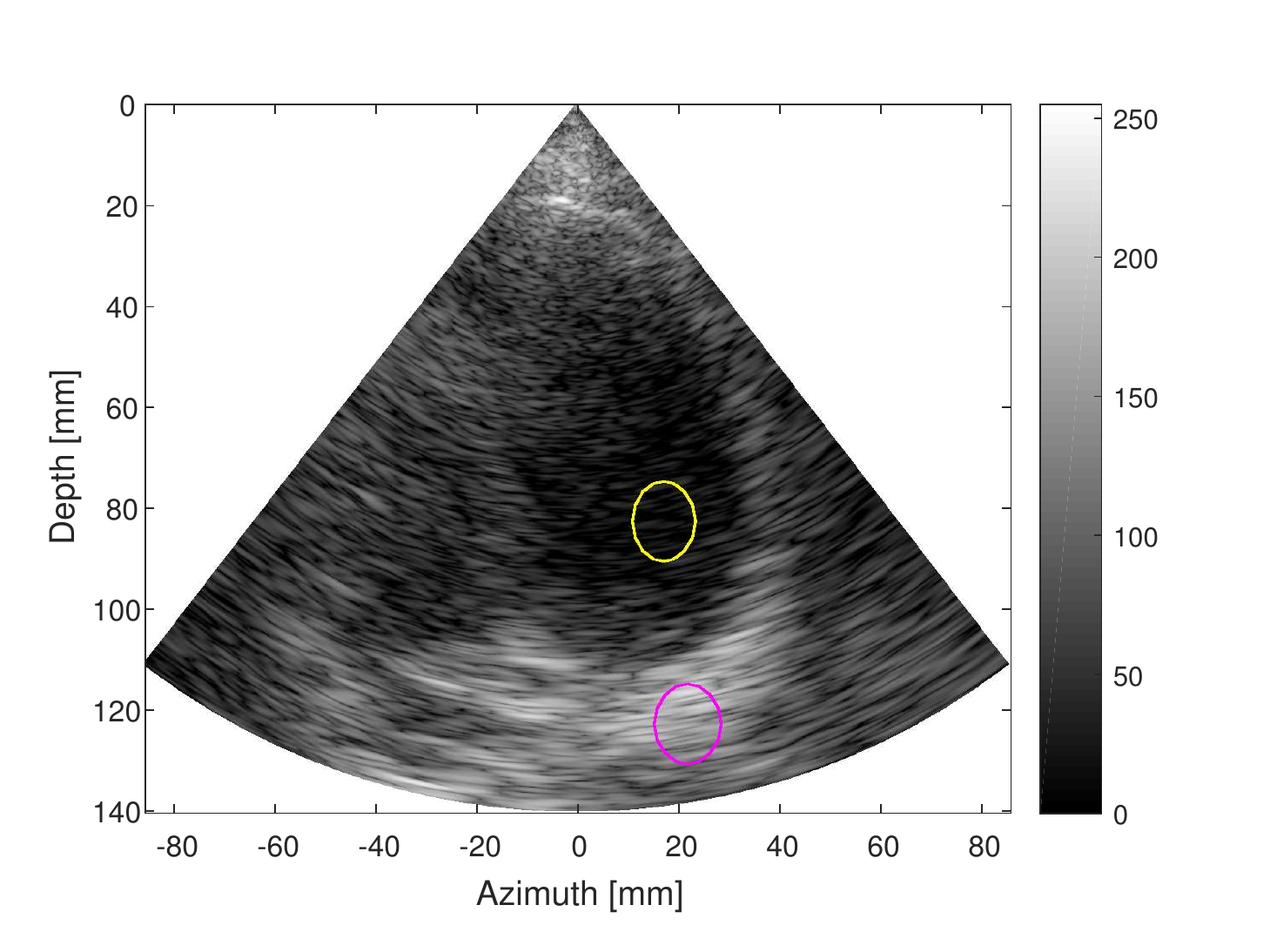} &
		\includegraphics[width = 0.33\textwidth]{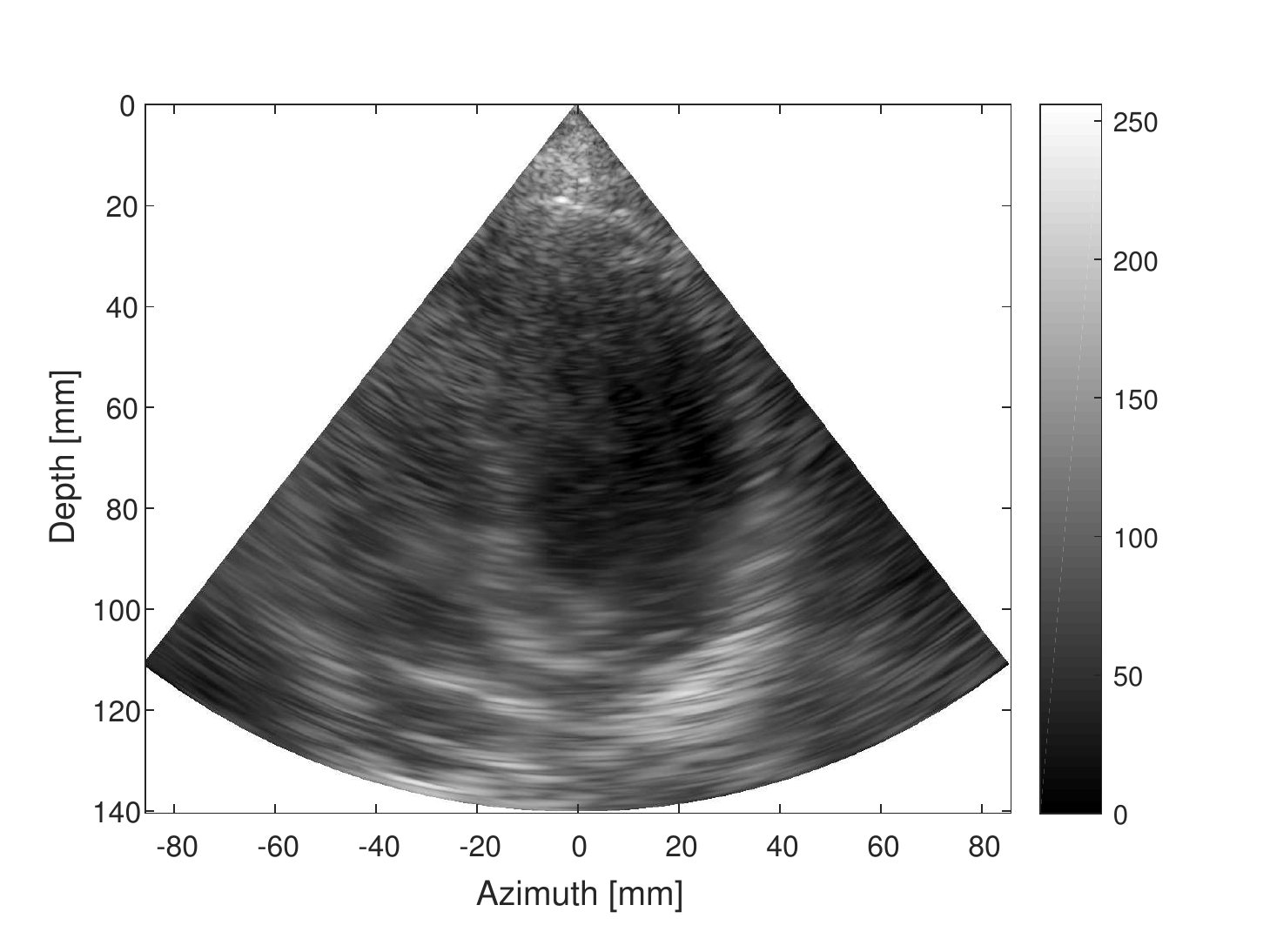} &
		\includegraphics[width = 0.33\textwidth]{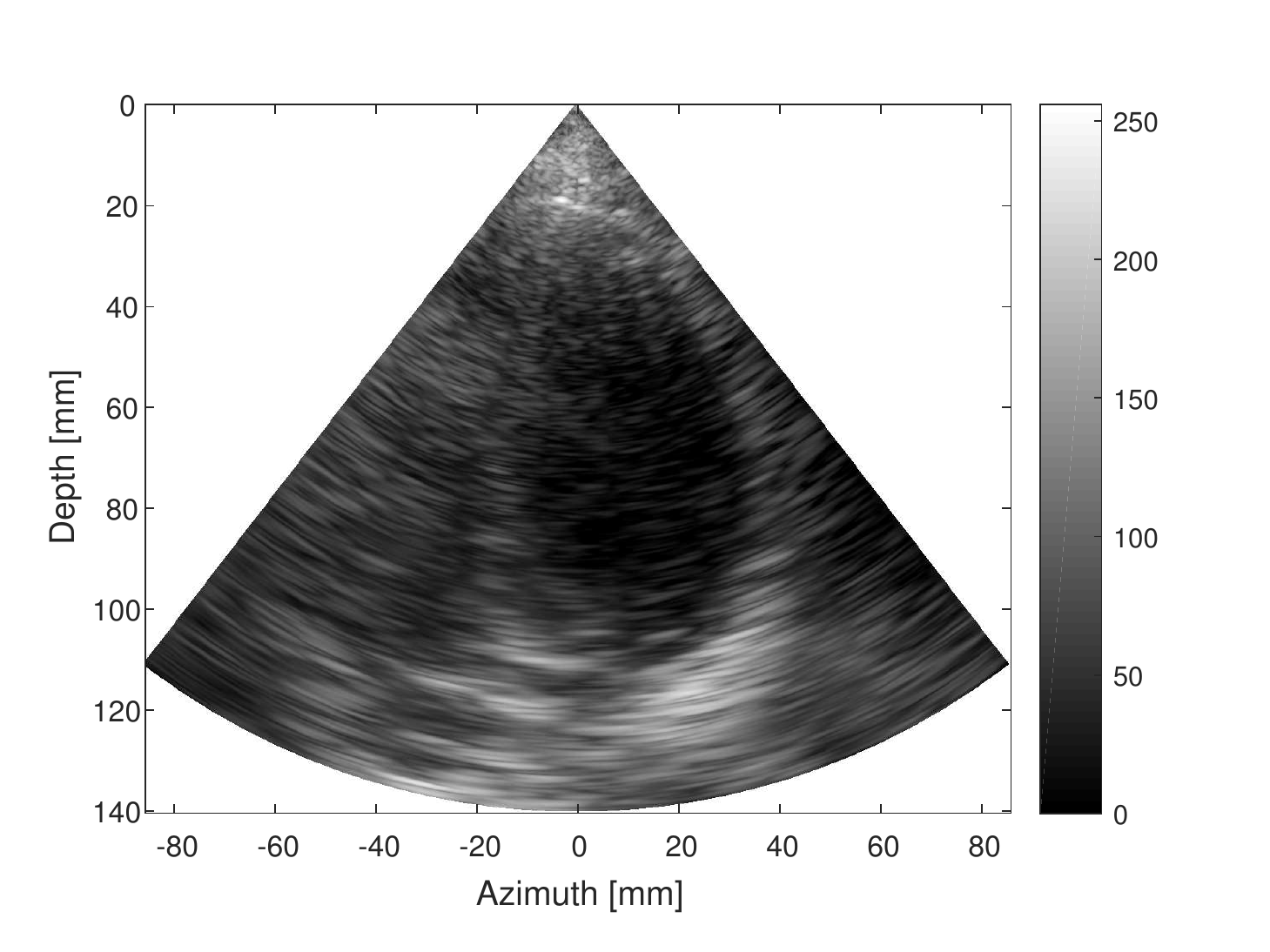}
	       \\
          (a) SLA   & (b) Learned Rx 20-MLA  & (c) Learned Tx-Rx 20-MLA \\
                      &
		\includegraphics[width = 0.33\textwidth]{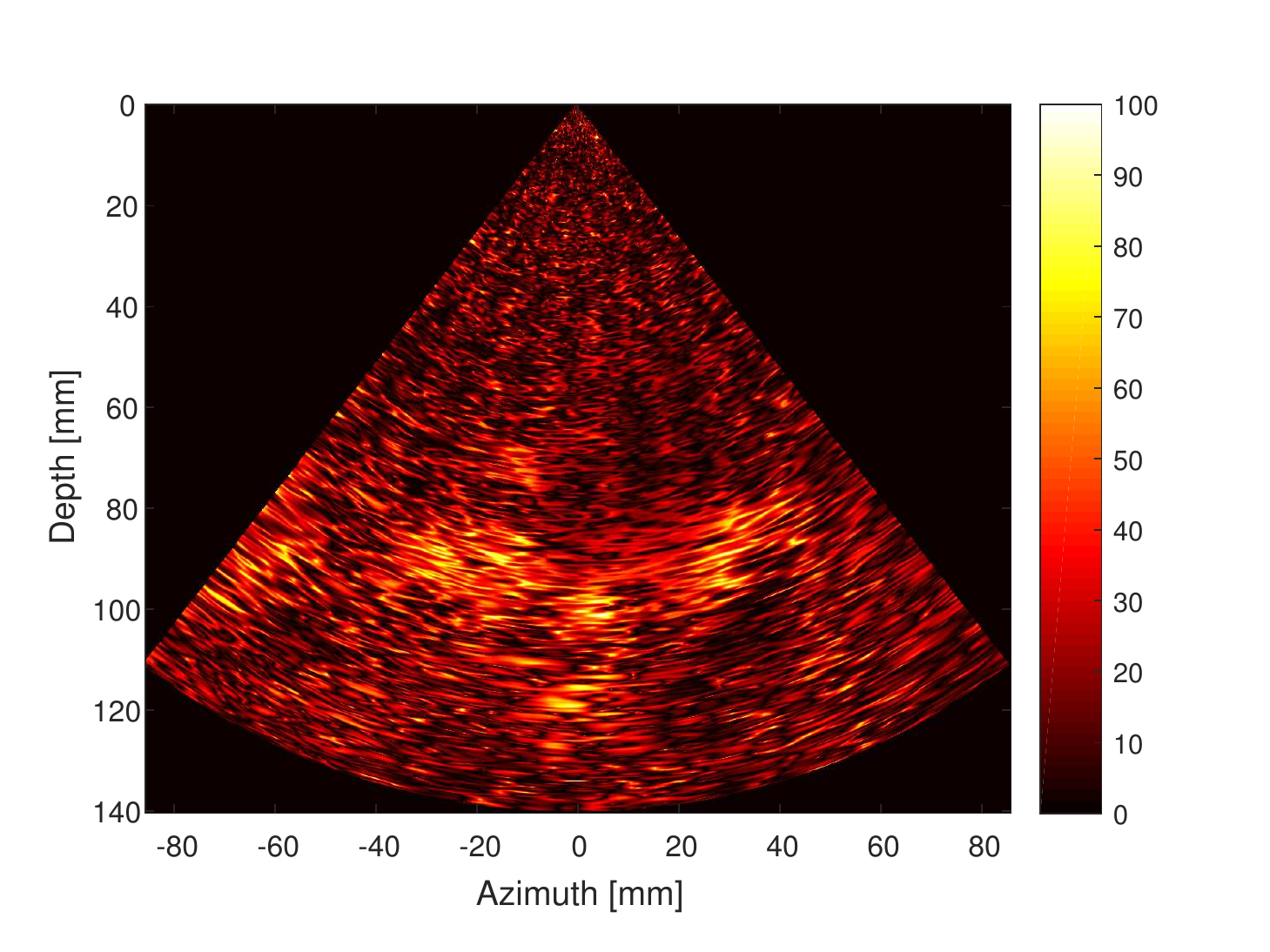} &
		\includegraphics[width = 0.33\textwidth]{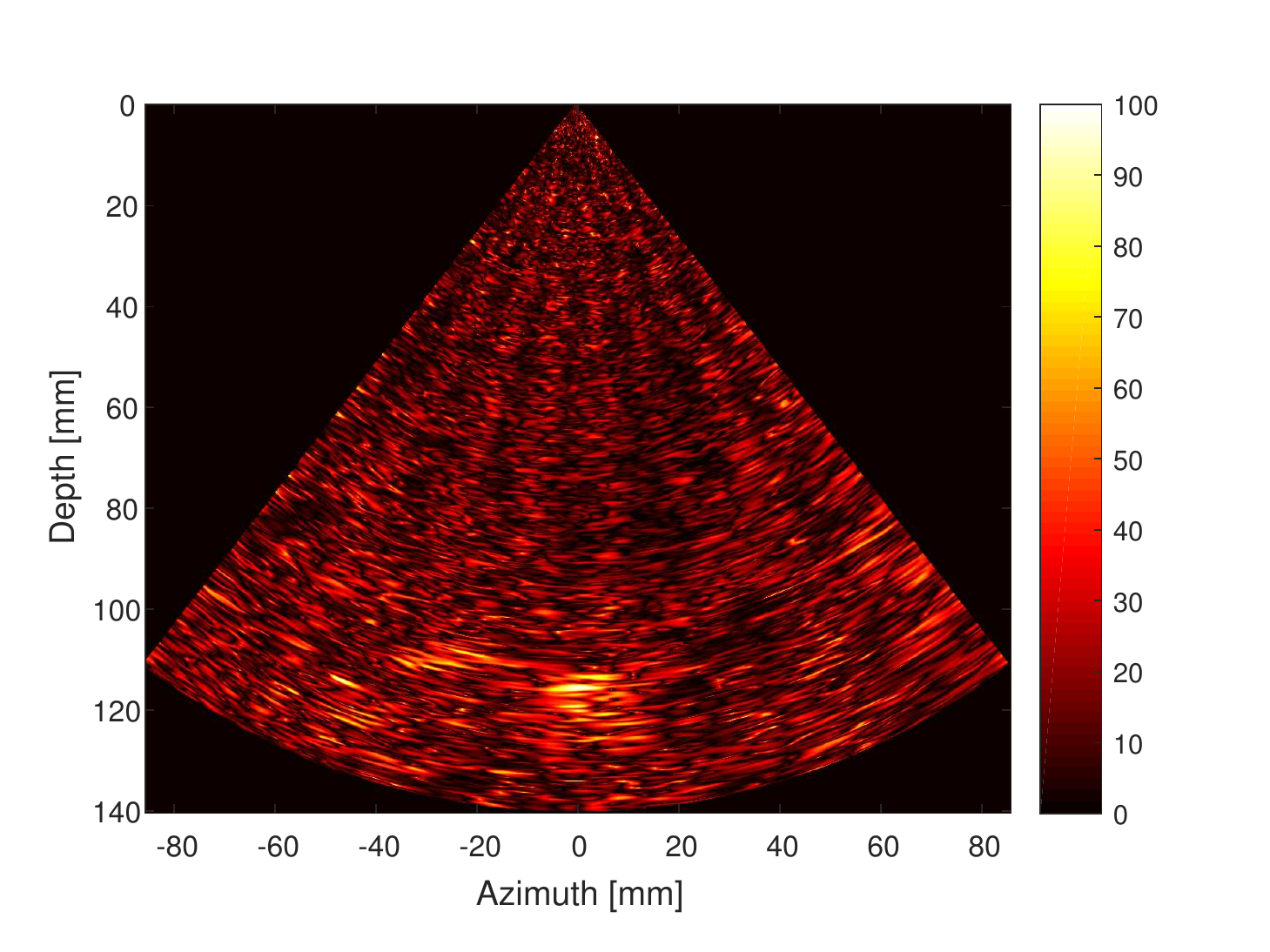} 
         \\
                      & difference((b), (a))  & difference((c), (a)) \\
	\end{tabular}   \\
  \end{minipage}
    \includegraphics[width=\textwidth]{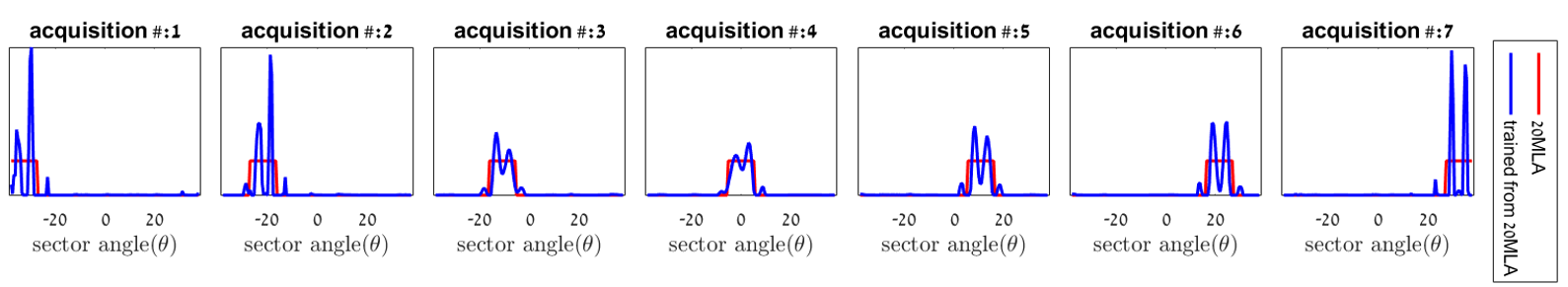}
	\caption{\small Visual comparison of \emph{Learned-Rx} and \emph{Learned Tx-Rx} settings of $20-$MLA on a test frame. The first row depicts (a) the ground truth SLA image, (b) reconstruction obtained from the \emph{Learned Rx} setting and (c) the reconstruction obtained from the \emph{Learned Tx-Rx} setting. The second row depicts the corresponding difference frames (with respect to the SLA image). The bottom row depicts the initial (red) and learned beampatterns (blue) of the $7$ acquisitions in the $20-$MLA setting.}
	\label{fig:example_in_paper}
\end{figure}

Visual inspection of the results of the two-stage training experiment for both different rates and different initializations settings, on one of the test frames is displayed in Figures \ref{fig:diff_MLA_rates}, \ref{fig:diff_inits} in the Appendix, along with the corresponding difference images (compared to SLA) and contrast (Cr), and contrast-to-noise (CNR) ratios (Tables \ref{tab:constrast_tables_diff_inits}, \ref{tab:constrast_tables_diff_rates}). These results suggest a better interpretability of the images generated from the jointly trained Tx-Rx models, especially for higher decimation rates (as displayed for the $20-$MLA initialization in Figure \ref{fig:example_in_paper}) and the less-directed initalizations (MLT and random). 

\paragraph{Generalization to phantom dataset. }
A phantom dataset consisting of $46$ frames was acquired with the same acquisition setup as of the cardiac dataset from a tissue mimicking phantom (GAMMEX Ultrasound $403$GS LE Grey Scale Precision Phantom). In order to evaluate the generalization performance of the proposed approach, we test all the networks that were originally trained on the cardiac samples on the phantom dataset. Results in Table \ref{tab:quantitaive_analysis_phantom} suggest that the proposed methodology while being trained on the cardiac data, generalizes well to the phantom, which is also consistent with the observations we made in our previous works \cite{vedula2018high, senouf2018high}. Firstly, this indicates that our reconstruction CNN does not overfit to the anatomy it was trained on. Secondly, and more interestingly, we can observe that the \textit{Learned Tx-Rx} setting consistently outperforms the \textit{Learned Rx} setting, which indicates that the transmit patterns learned over the cardiac data also transfer well to the phantoms.

\begin{table}[!htbp]

\centering
\setlength\tabcolsep{1.5pt} 
\begin{tabular}{l||lll|lll|lll}

              &      & 7-MLA &          &      & 10-MLA &          &      & 20-MLA &           \\ 
\hline
              & PSNR     & SSIM    & L1-error    & PSNR     & SSIM     & L1-error     & PSNR     & SSIM      & L1-error  \\
Learned Rx    & 40.92    & $0.978$   & $5.3$       & 32.09    & 0.911    & 9.91        & 30.23   & 0.901     & 12.94     \\
Learned Tx-Rx & 43.73     & $0.989$  & $3.35$       & 31.14    & 0.92     & 7.62        & 31.92    & 0.903      & 10.53 
\vspace{0.5cm}
\end{tabular}

\begin{tabular}{l||lll|lll|lll}
              &      & 10-MLA &          &      & 10-MLT &          &      & 10-random &           \\ 
\hline
              & PSNR   & SSIM  & L1-error & PSNR   & SSIM   & L1-error & PSNR  & SSIM   & L1-error  \\
Learned Rx    & 32.09    & 0.911    & 9.91   & 31.05  & 0.624   & 15.91   & 31  & 0.667  & 16.34        \\
Learned Tx-Rx & 31.14    & 0.92     & 7.62    & 32.098  & 0.711   & 13.765       & 31 & 0.76   & 14.276       
\end{tabular}
\caption{Generalization to phantom dataset. Comparison of average PSNR, SSIM and $L_1$ error measures between different decimation rates of transmissions (top) and different initializations (bottom). Top and bottom rows indicate the results corresponding to learned Rx and learned Tx-Rx experiment settings, respectively.}
\label{tab:quantitaive_analysis_phantom}
\end{table}

\paragraph{Learned beam patterns.}
A visualization of the learned beam profiles for $7-$, $10-$ and $20-$MLA initializations as presented in the Appendix in Figures \ref{fig:BeamPlot7MLA}, \ref{fig:BeamPlot10MLA} and \ref{fig:BeamPlot20MLA}, respectively. These profiles suggest that the general trend of the beam transformation is towards higher directivity. The wider the initialized beams are (higher MLA rates), the greater is the increase in the directivity, such that for the very wide $20-$MLA initialization (as depicted in Figure \ref{fig:example_in_paper}), the beam pattern converges into two splitted narrower beams. 
The visualization of the beam profiles of the $10-$ MLT and $10-$ random initializations, as displayed in the Appendix in Figures \ref{fig:BeamPlot10MLT}  and \ref{fig:BeamPlot10random}, respectively, suggest that there is a trade-off between the directivity of the beam and the field of view it covers. The $10-$MLT profile displays a trend towards widening the simultaneously transmitted narrow beams, whereas for the random initialization, some of the beams stays un-directed and some of them approach the MLT pattern.

\section{Conclusion and future directions}
We have demonstrated, as  a proof-of-concept, that jointly learning the transmit patterns with the receive beamforming  provides greater improvements to the image quality. 
It should be mentioned that since the beam patterns trained from the MLA initialization displayed the optimal results, we can assume the models have not reached the globally optimal configuration -- otherwise, all patterns would have converged to similar performance. This calls for better optimization techniques which are more robust to initialization in regression problems in general and in imaging in particular. It should be noted that in all the experiments mentioned within this paper, delay-and-sum beamformed SLA was considered as the ground truth reference to the neural network. However, the presented methodology can be simply extended to more sophisticated beamformers such as minimum-variance beamforming by modifying the reference envelope ultrasound image appropriately \cite{simson2018Deepformer}, or to other tasks such as estimating the speed-of-sound \cite{Feigin2018DLSoS} or the scatterer maps of the tissues \cite{vedula2017towards}. It would be particularly interesting to explore such learning-based beam pattern designs to combat the frame-rate vs. resolution tradeoffs in the case of 2D ultrasound probes and to enable efficient computational sonography \cite{mateus2018compsonography}.

An interesting insight observed from the $10-$random experiment is that the learned beam profiles perform significantly better than transmitting random undirected beam patterns both with the delay-and-sum and the learned beamformers. This makes us wonder whether transmitting planar waves is really optimal with a learned receive pipeline. Lastly, in the proposed work, the learned transmit patterns are fixed during post-training. It would be interesting to explore how to design transmit protocols, that are scene or anatomy adaptive, and extend the proposed methodology to the non-linear second-harmonic imaging. We believe that all these directions would initiate a new line of research towards building efficient learning-driven ultrasound imaging.

\midlacknowledgments{This research was funded by ERC StG RAPID. We thank Prof. Dan Adam for making his GE machine available to us. }

\bibliography{midl-samplepaper}

\appendix

\begin{figure}
    \centering
    \includegraphics[width=0.80\textwidth]{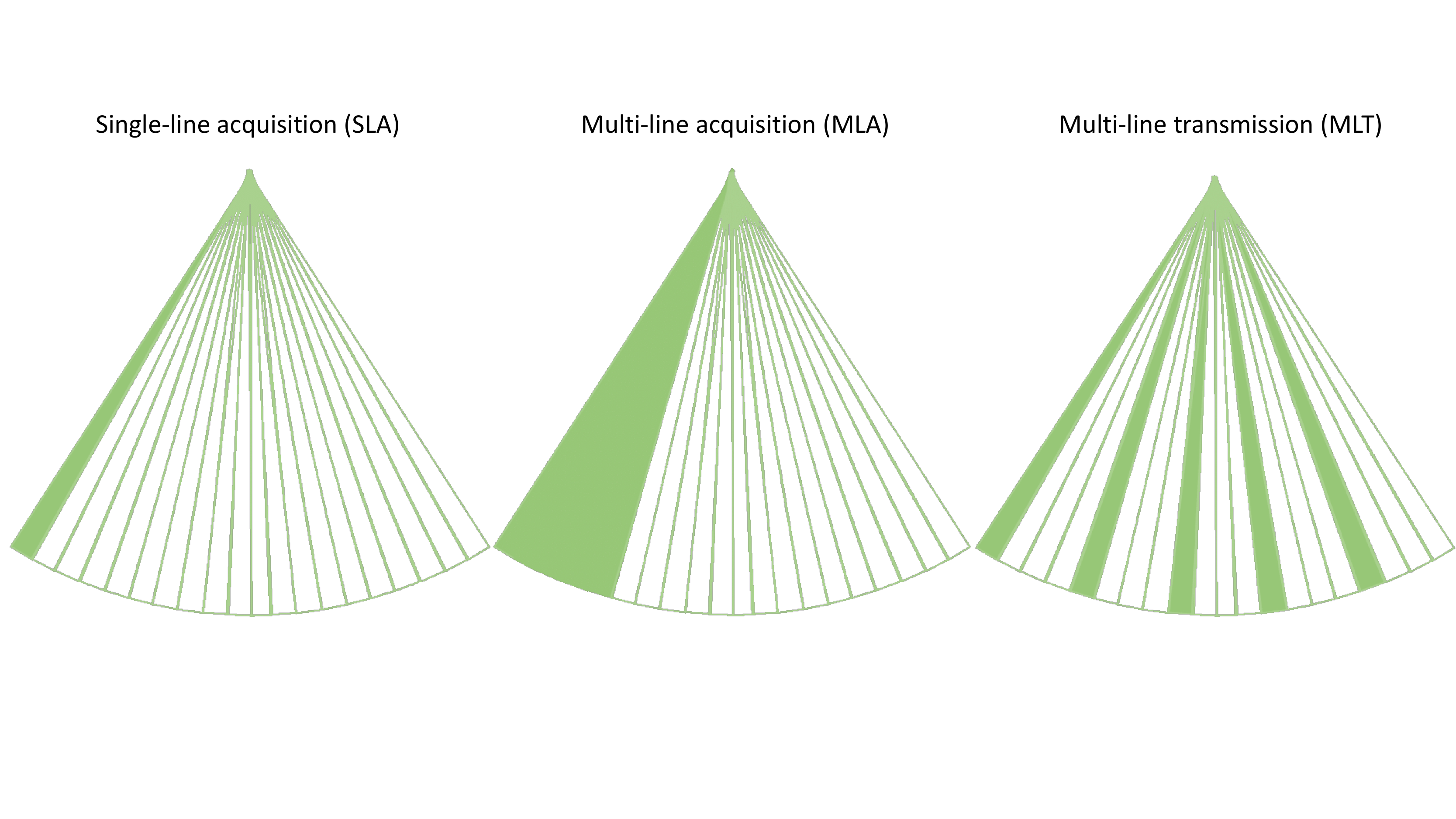}
    \caption{SLA/SLT vs. MLA, MLT}
    \label{fig:MLAMLT}
\end{figure}

\begin{sidewaysfigure}[h]
    \centering
    \includegraphics[width=1.1\textheight, height=0.8\textwidth]{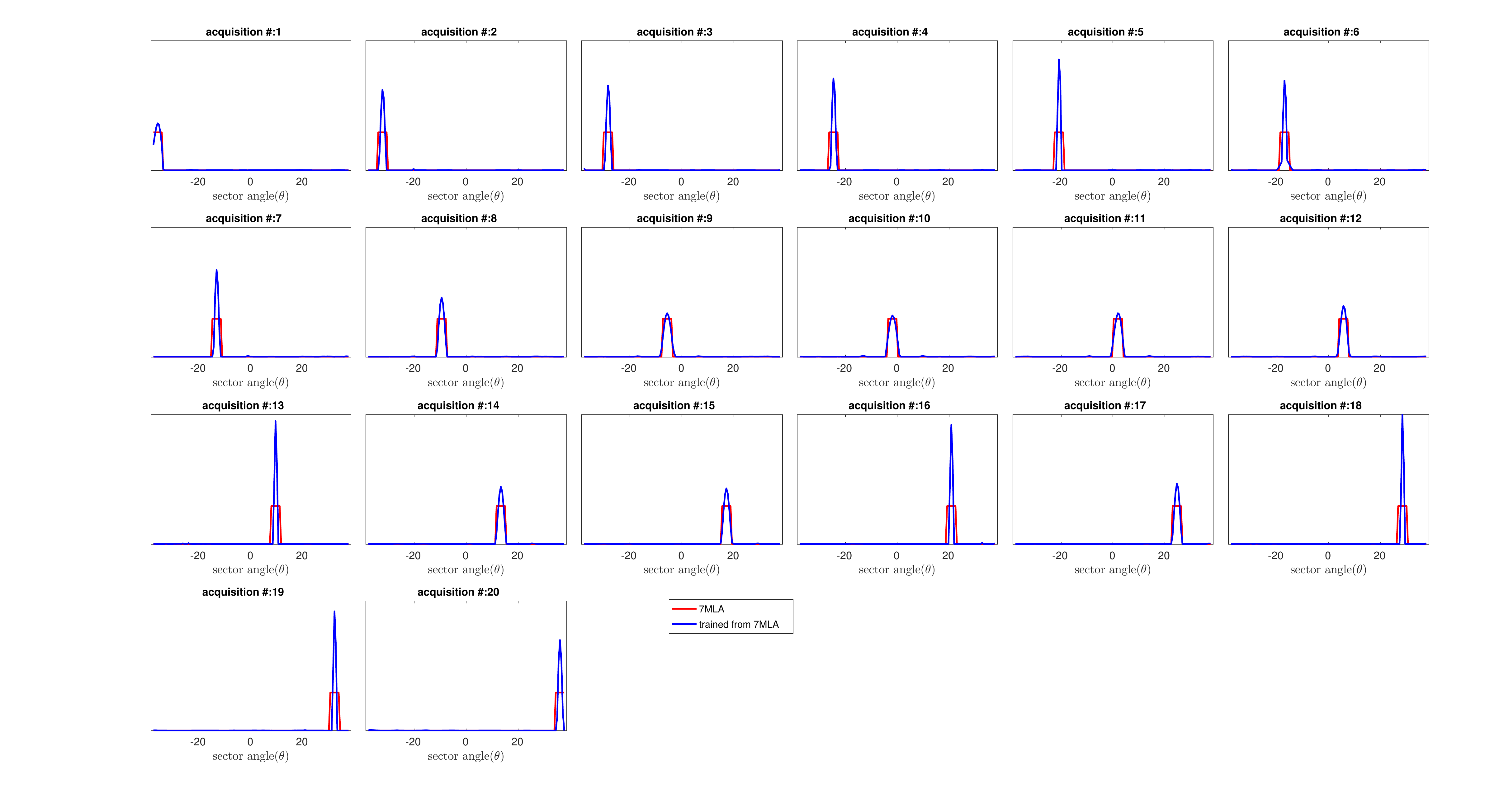}
    \caption{7-MLA: fixed vs. learned transmission}
    \label{fig:BeamPlot7MLA}
\end{sidewaysfigure}

\begin{sidewaysfigure}[h]
    \centering
    \includegraphics[width=1.1\textheight, height=0.8\textwidth]{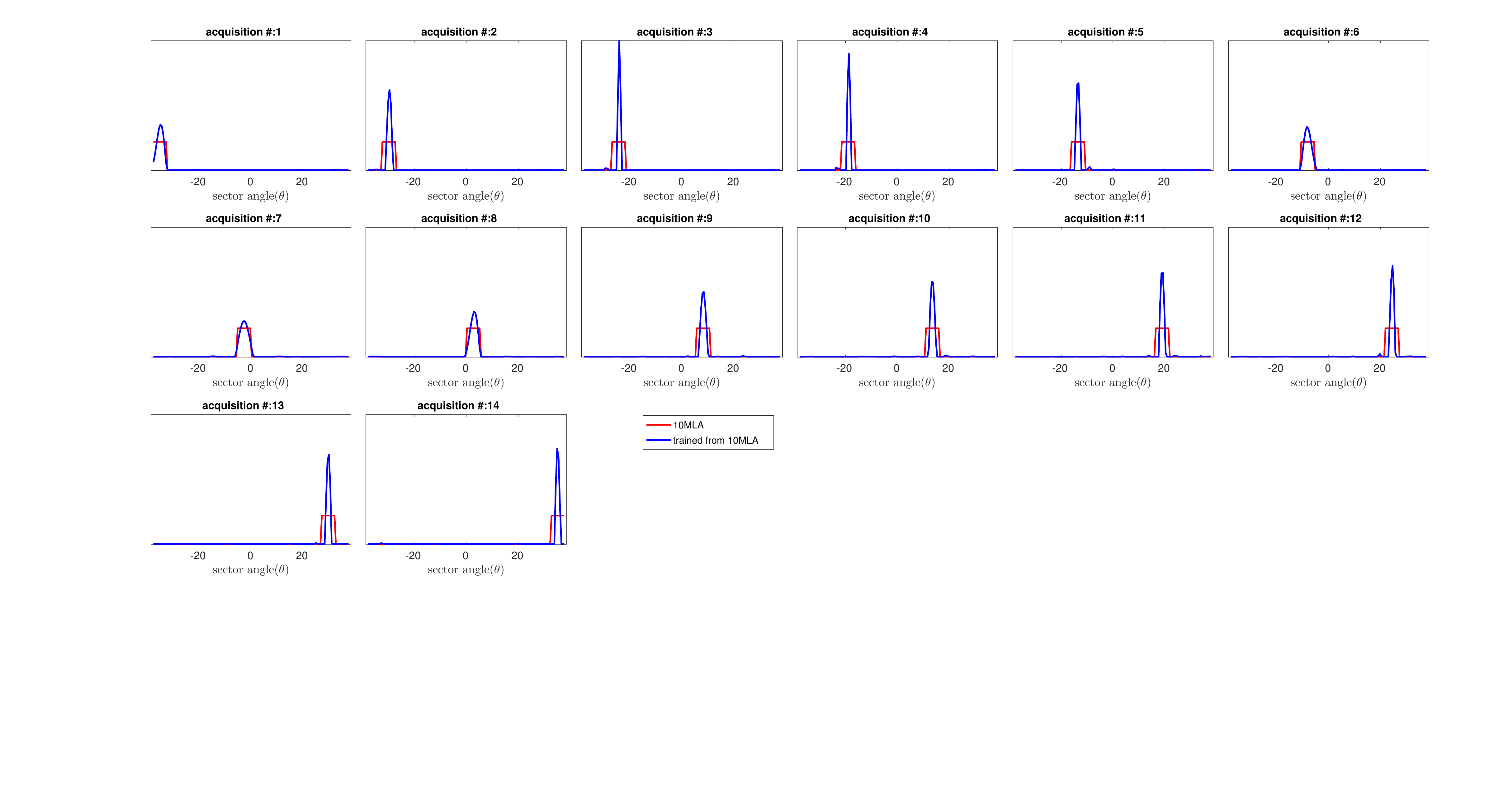}
    \caption{10-MLA: fixed vs. learned transmission}
    \label{fig:BeamPlot10MLA}
\end{sidewaysfigure}

\begin{sidewaysfigure}[h]
    \centering
    \includegraphics[width=1.1\textheight, height=0.8\textwidth]{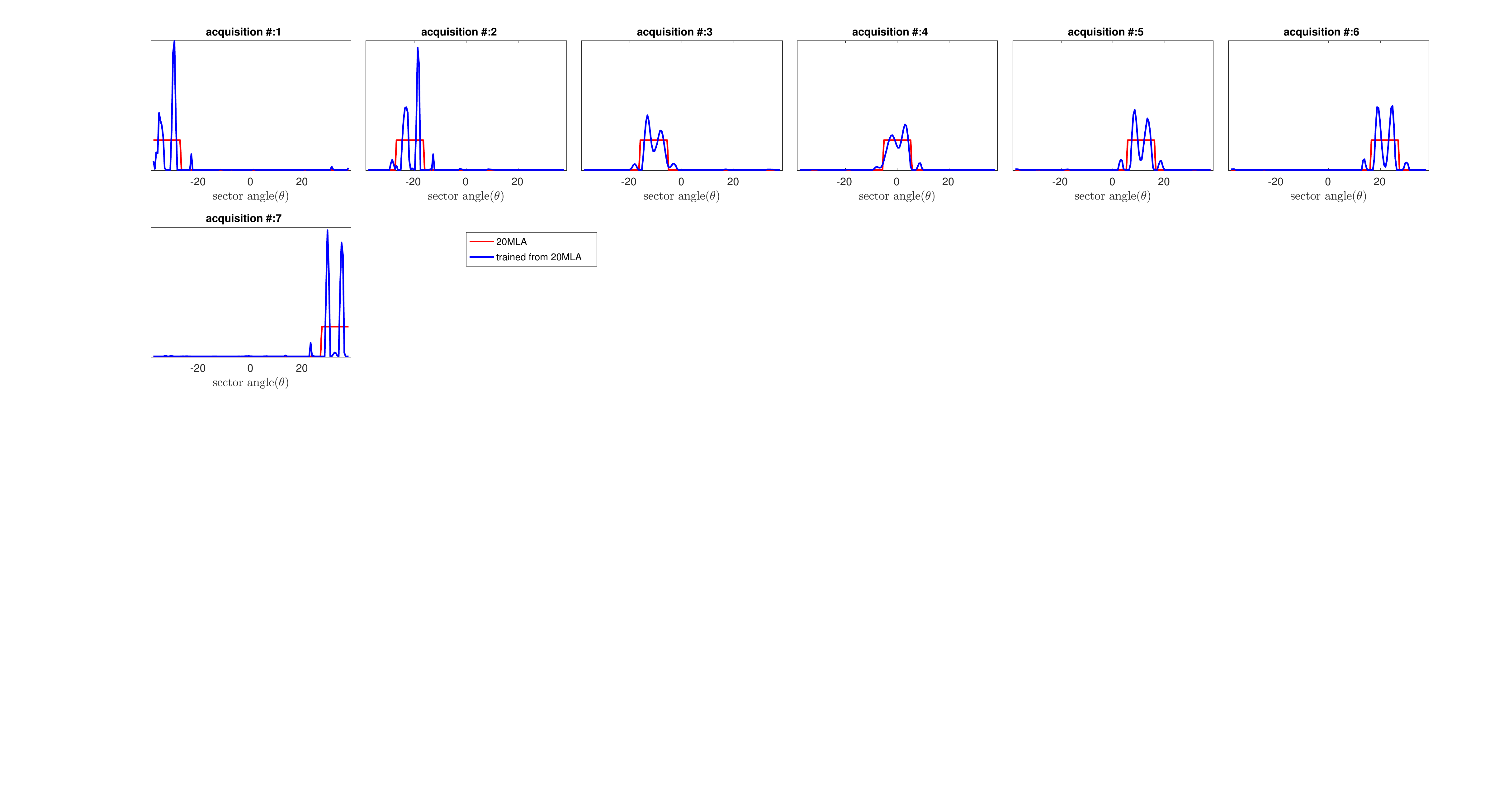}
    \caption{20-MLA: fixed vs. learned transmission}
    \label{fig:BeamPlot20MLA}
\end{sidewaysfigure}

\begin{sidewaysfigure}[h]
    \centering
    \includegraphics[width=1.1\textheight, height=0.8\textwidth]{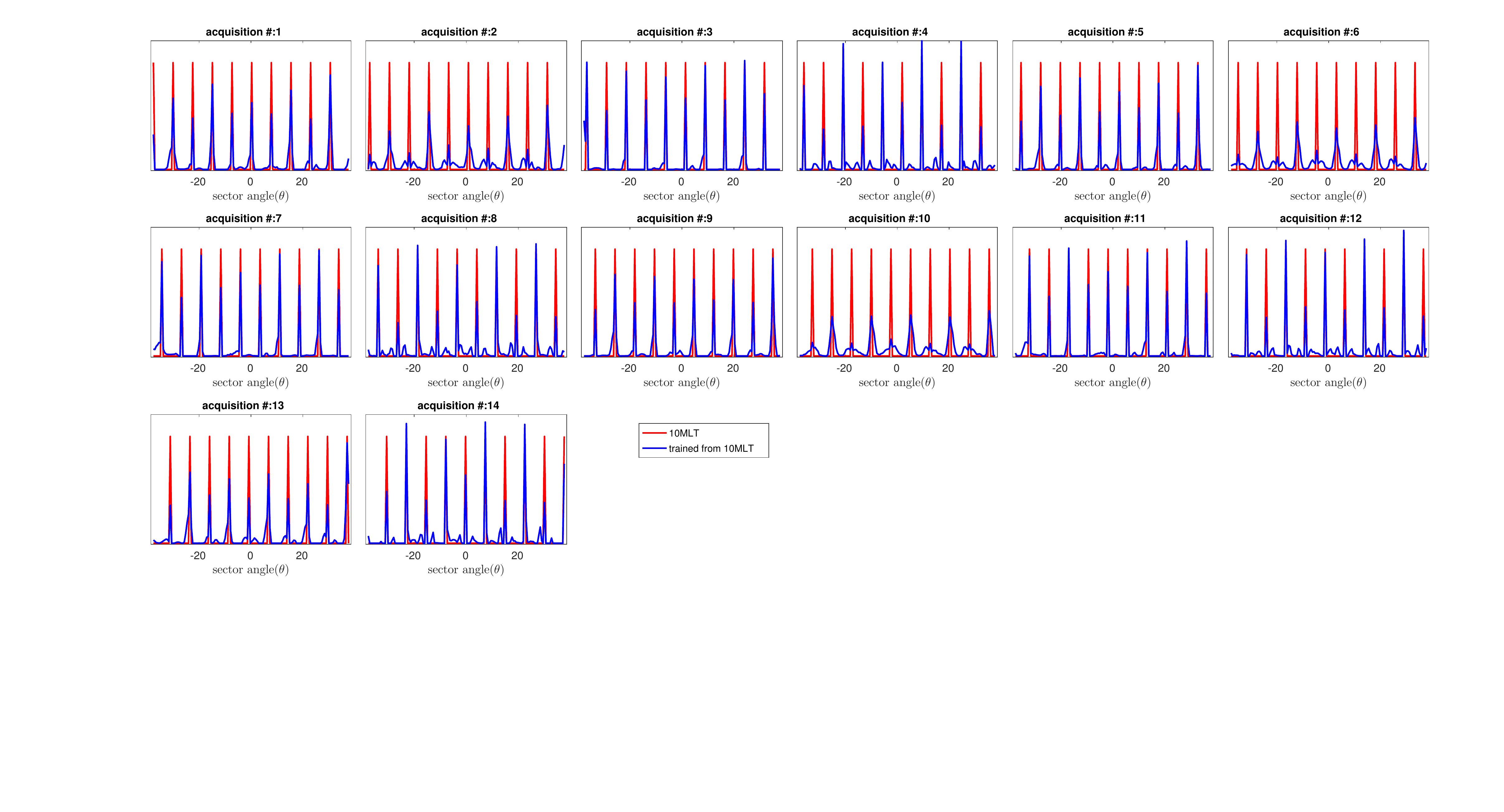}
    \caption{10-MLT: fixed vs. learned transmission}
    \label{fig:BeamPlot10MLT}
\end{sidewaysfigure}

\begin{sidewaysfigure}[h]
    \centering
    \includegraphics[width=1.1\textheight, height=0.8\textwidth]{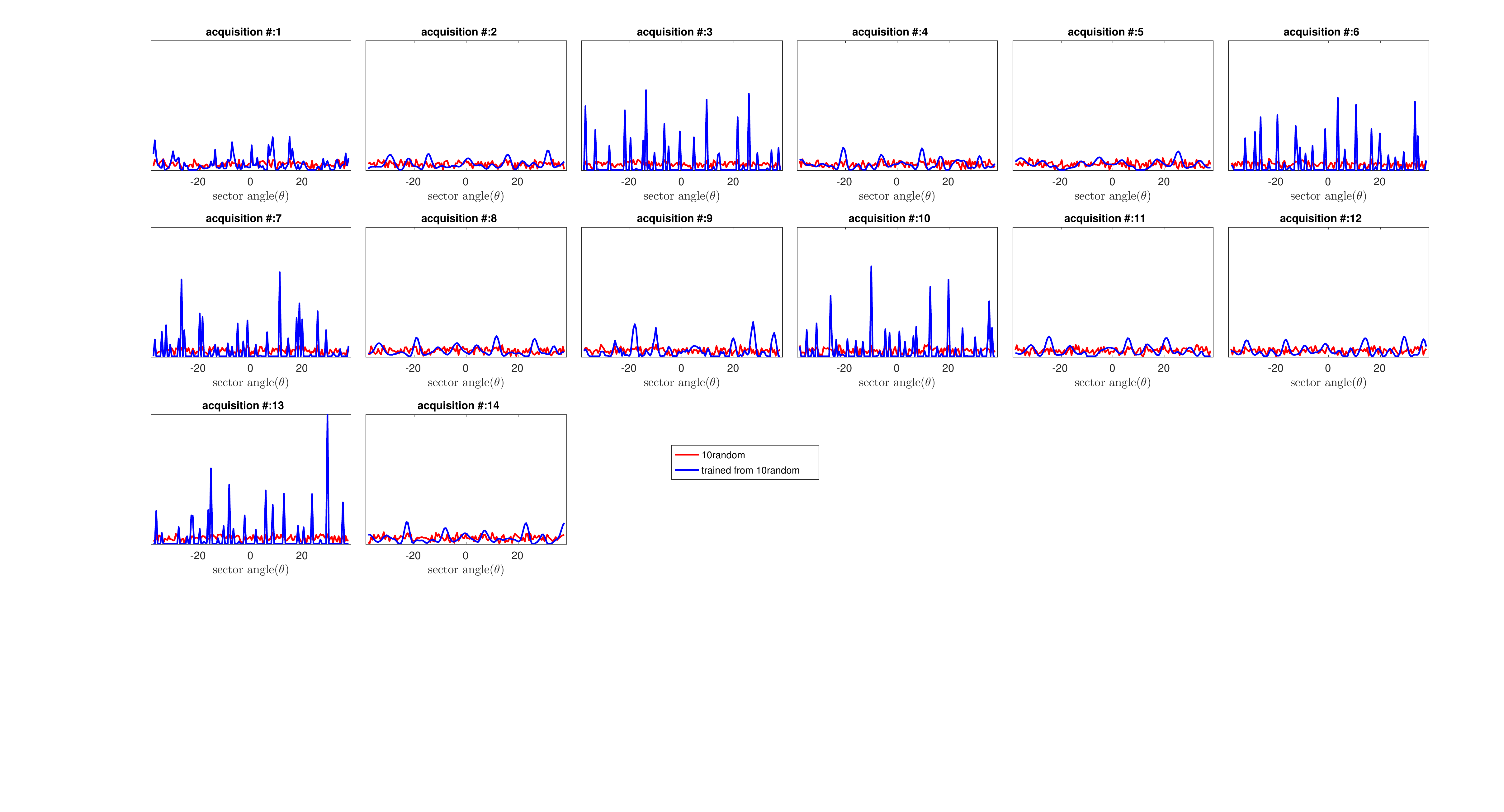}
    \caption{10-random: fixed vs. learned transmission}
    \label{fig:BeamPlot10random}
\end{sidewaysfigure}

\begin{sidewaysfigure}[ht]
\begin{minipage}[]{\linewidth}
	\begin{tabular}{ c@{\hskip 0.001\textwidth}c@{\hskip 0.001\textwidth}c@{\hskip 0.001\textwidth}c@{\hskip 0.001\textwidth}c } 
		\includegraphics[width = 0.24\textwidth]{results/f2f11_SLA.pdf} &
		\includegraphics[width = 0.24\textwidth]{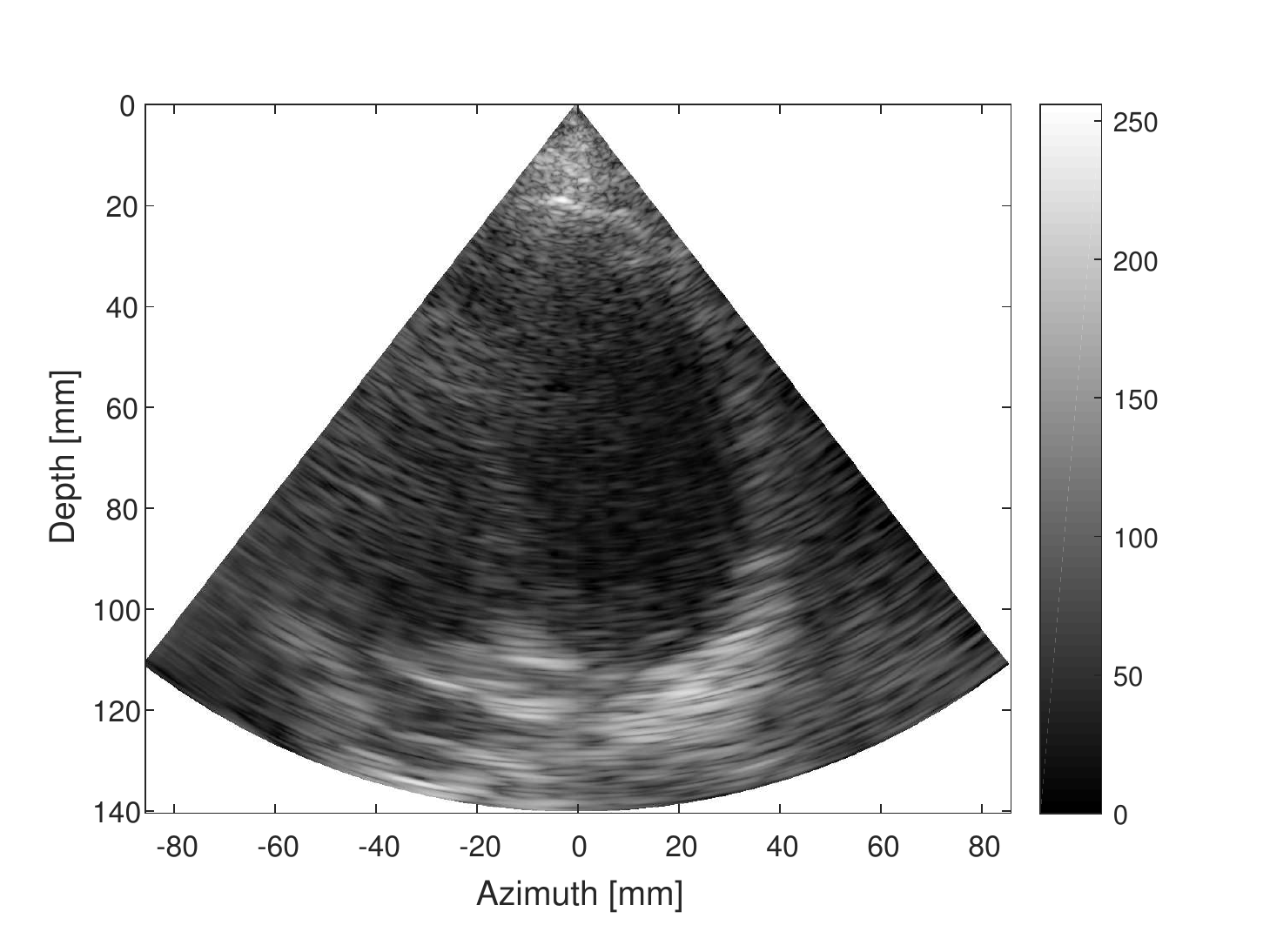} &
		\includegraphics[width = 0.24\textwidth]{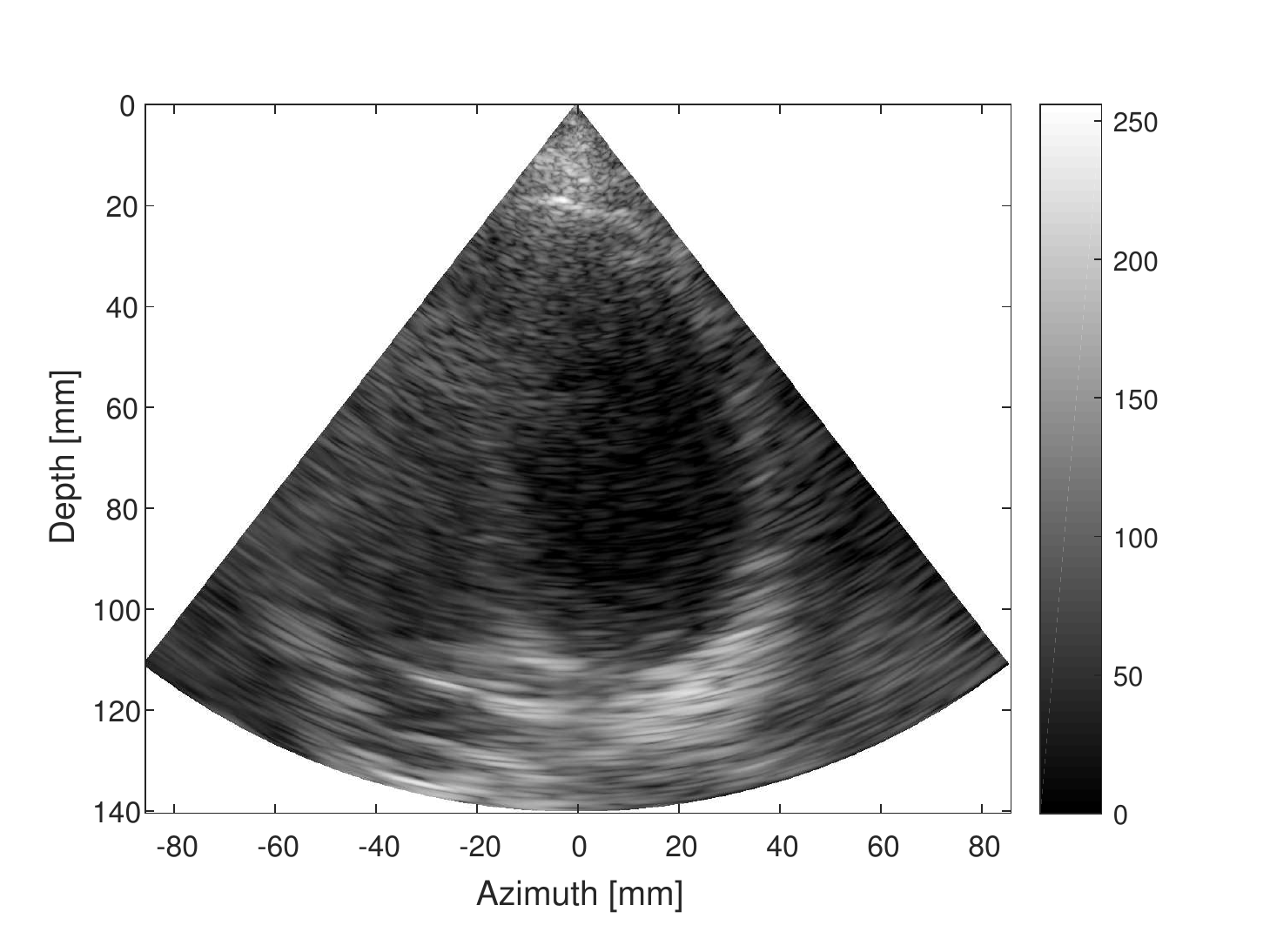} &
	    \includegraphics[width = 0.24\textwidth]{results/f2f11_fixed20MLA.pdf} 
	       \\
          1.1(a) SLA   & 1.1(b) Learned Rx 7-MLA  & 1.1(c) Learned Rx 10-MLA & 1.1(d) Learned Rx 20-MLA & \\
                     &
		\includegraphics[width = 0.24\textwidth]{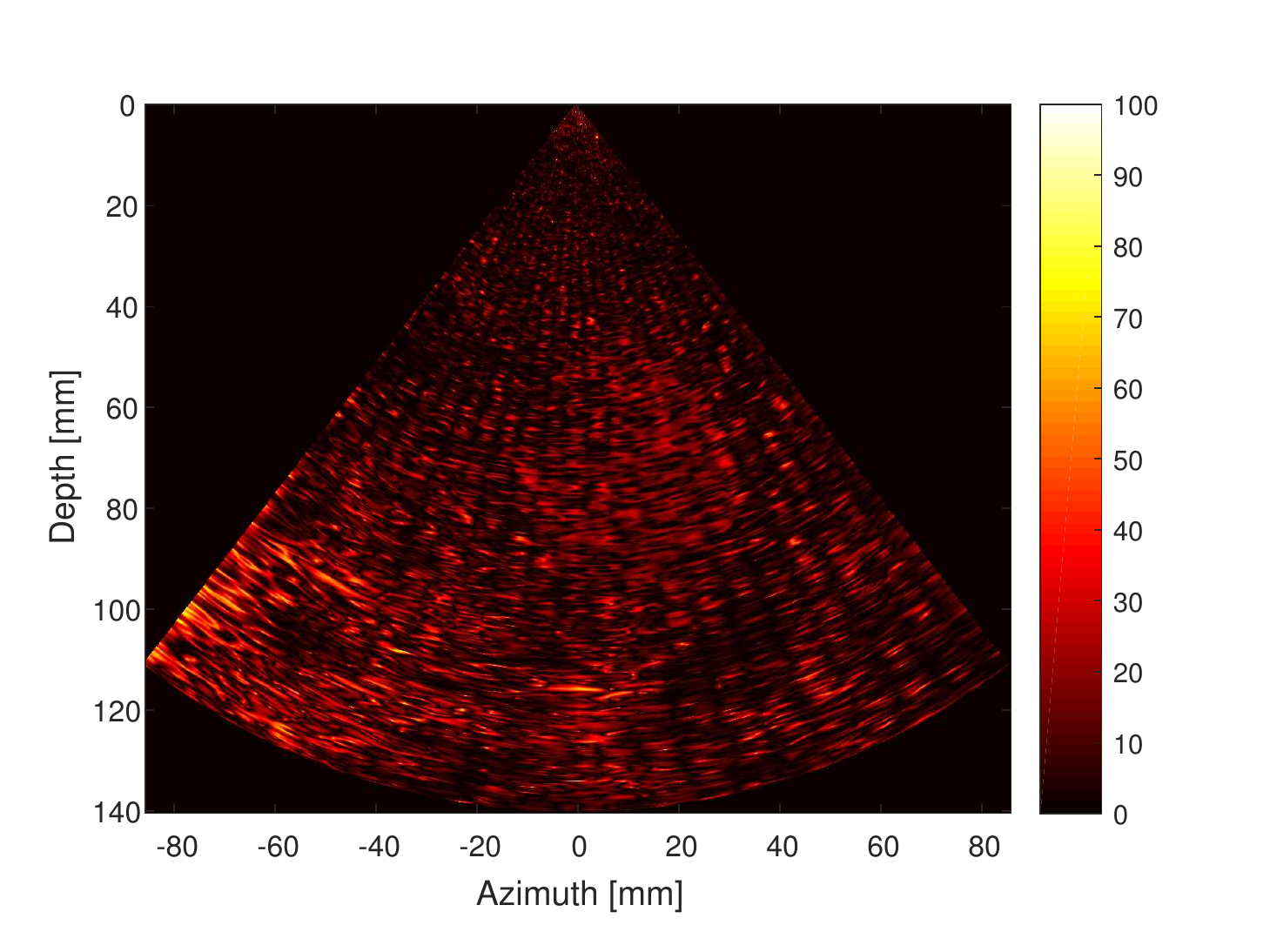} &
		\includegraphics[width = 0.24\textwidth]{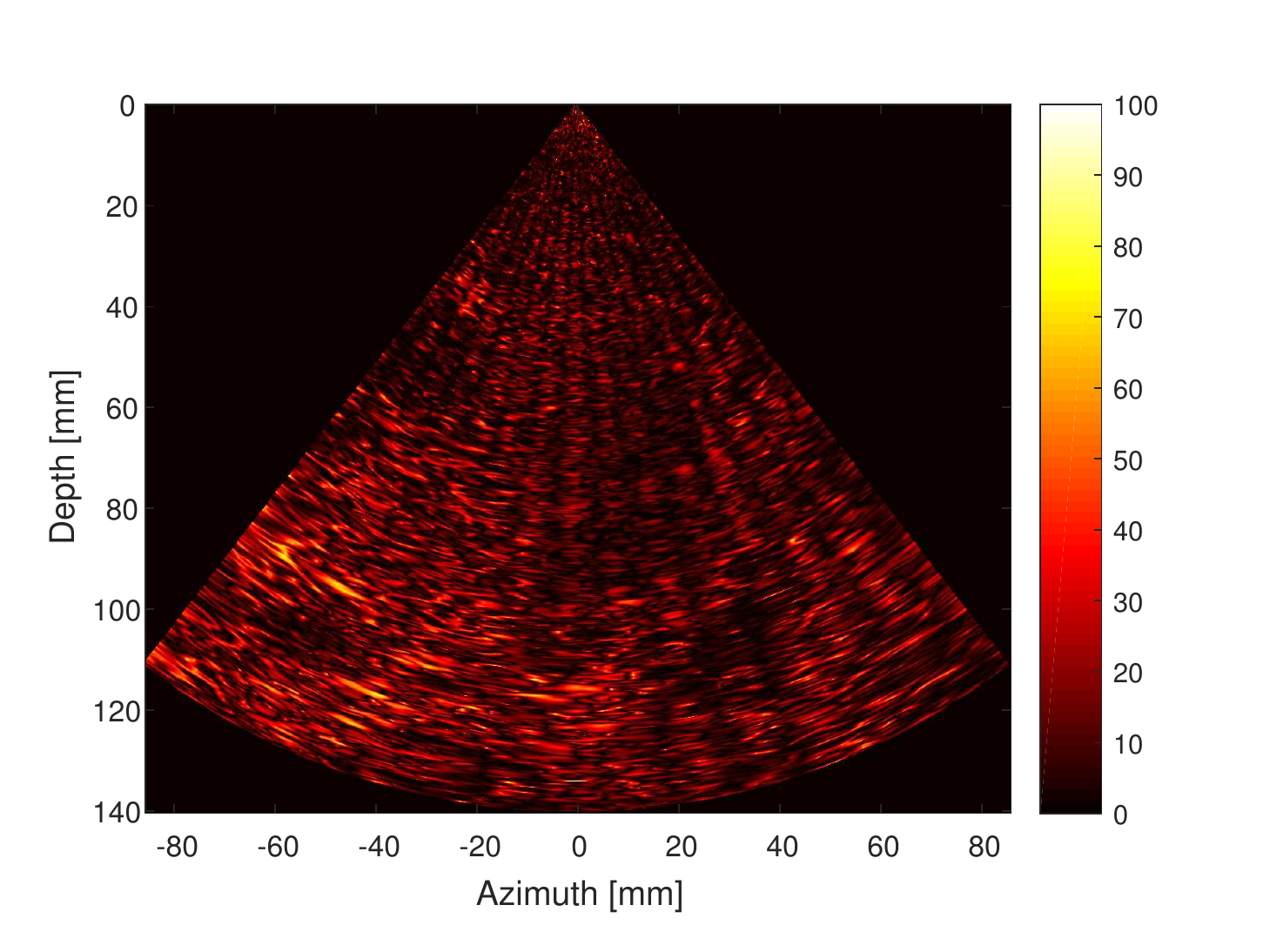} &
		\includegraphics[width = 0.24\textwidth]{results/f2f11_fixed20MLA_diff.pdf} 
         \\
                       & difference(1.1(b), 1.1(a))  & difference(1.1(c), 1.1(a)) &  difference(1.1(d), 1.1(a))& \\
           &
		\includegraphics[width = 0.24\textwidth]{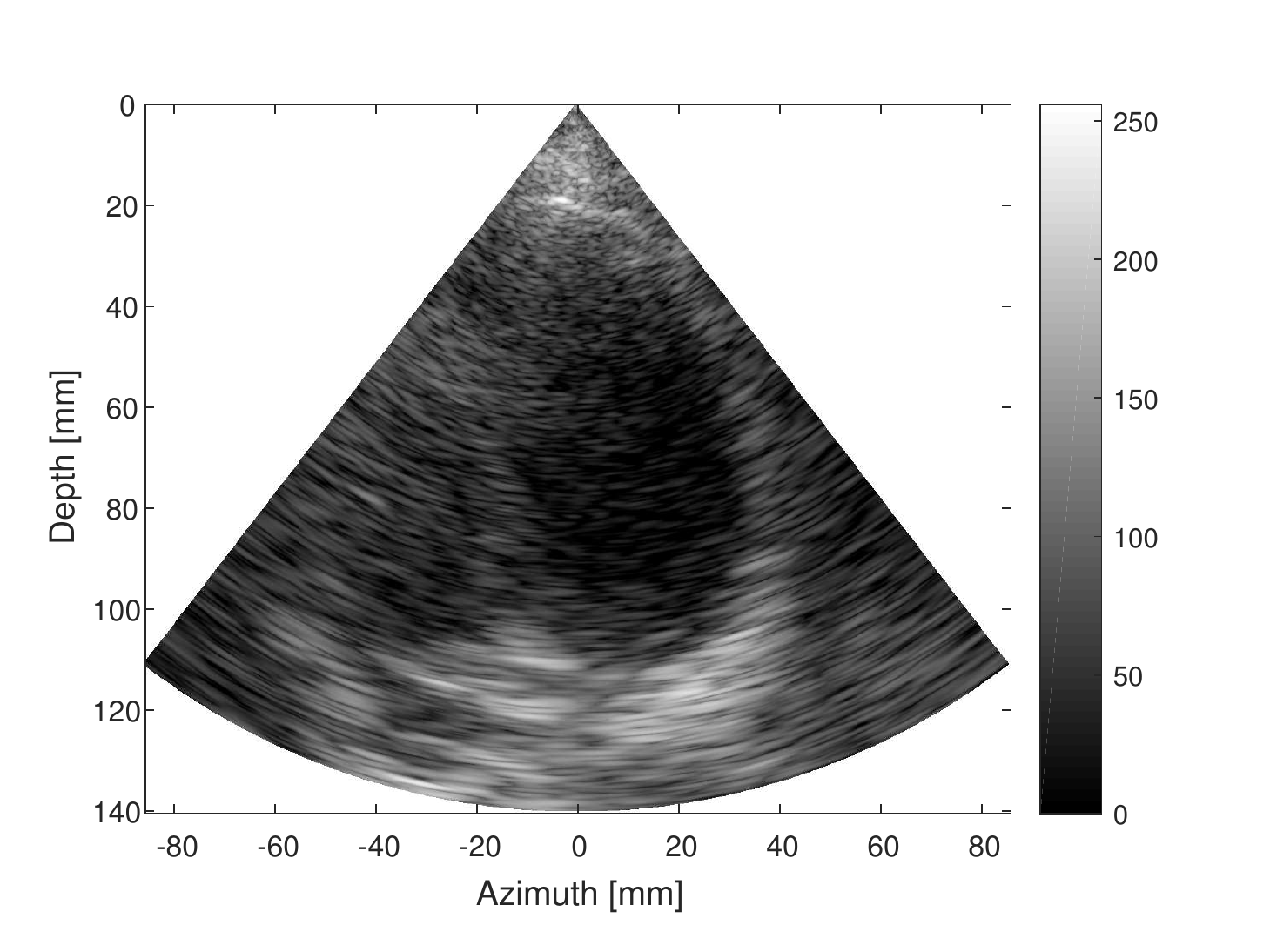} &
		\includegraphics[width = 0.24\textwidth]{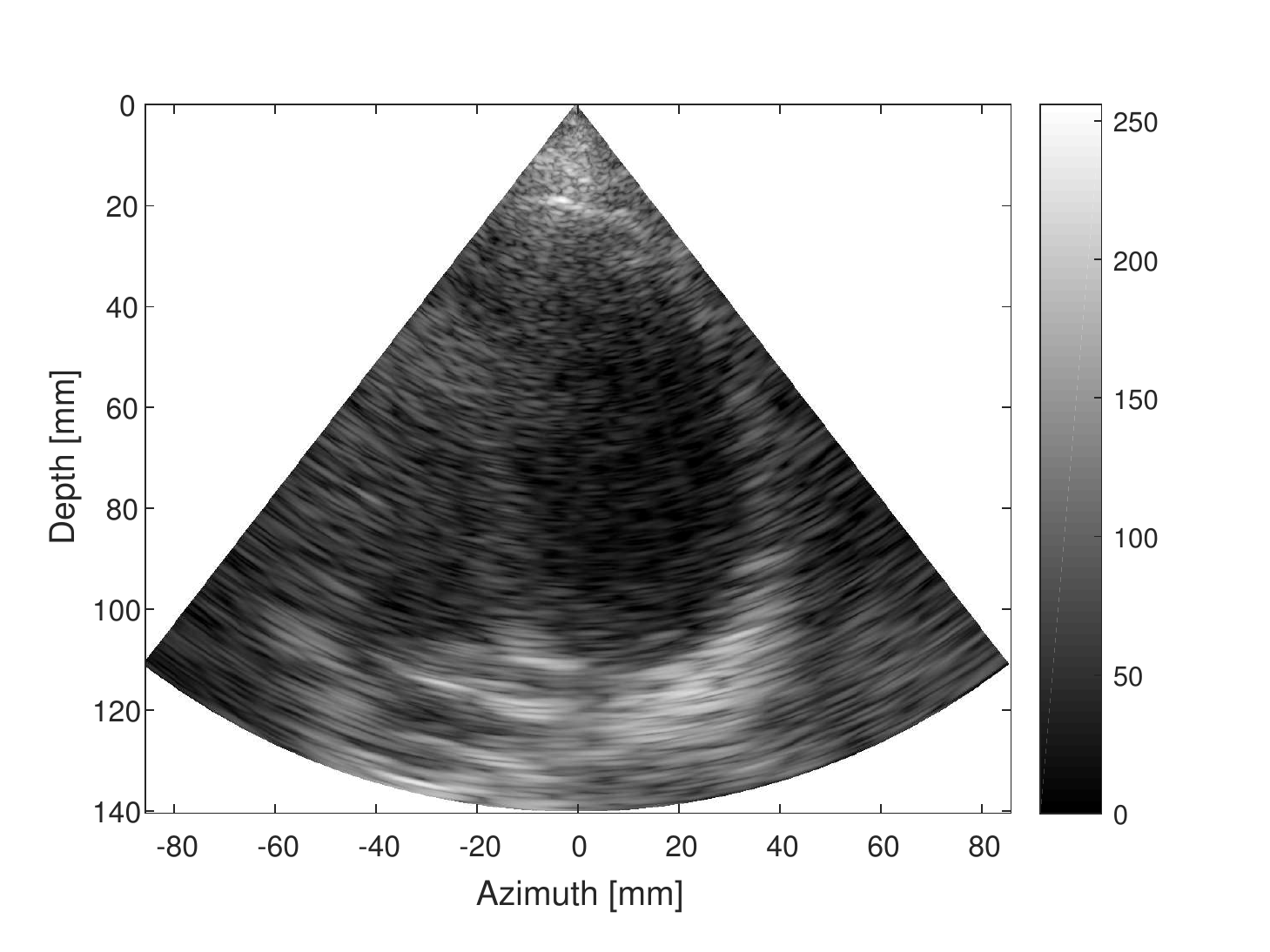} &
		\includegraphics[width = 0.24\textwidth]{results/f2f11_trained20MLA.pdf} 
         \\
              & 1.2(b) Learned Tx-Rx 7-MLA  & 1.2(c) Learned Tx-Rx 10-MLA & 1.2(d) Learned Tx-Rx 20-MLA & \\
                         &
		\includegraphics[width = 0.24\textwidth]{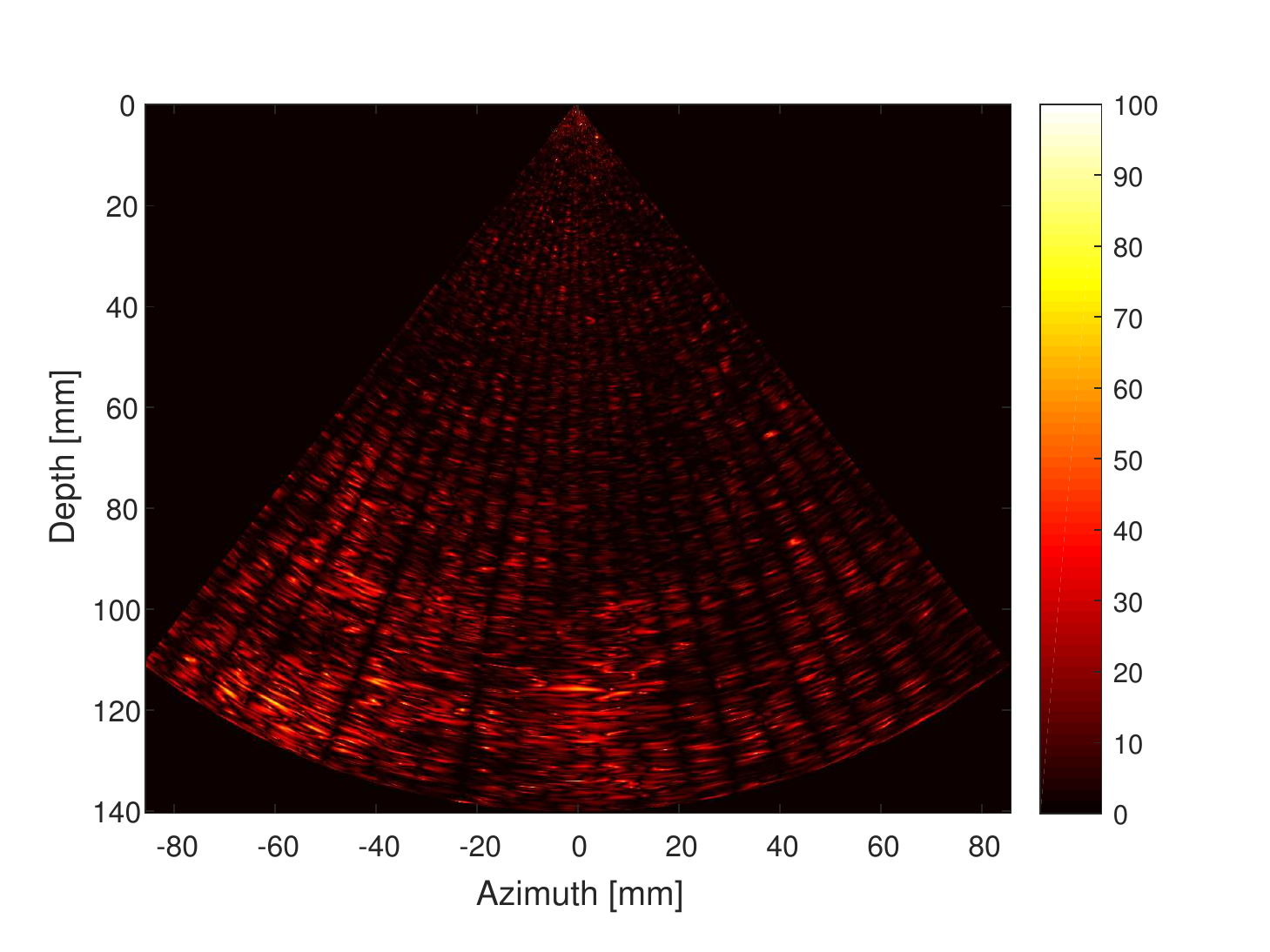} &
		\includegraphics[width = 0.24\textwidth]{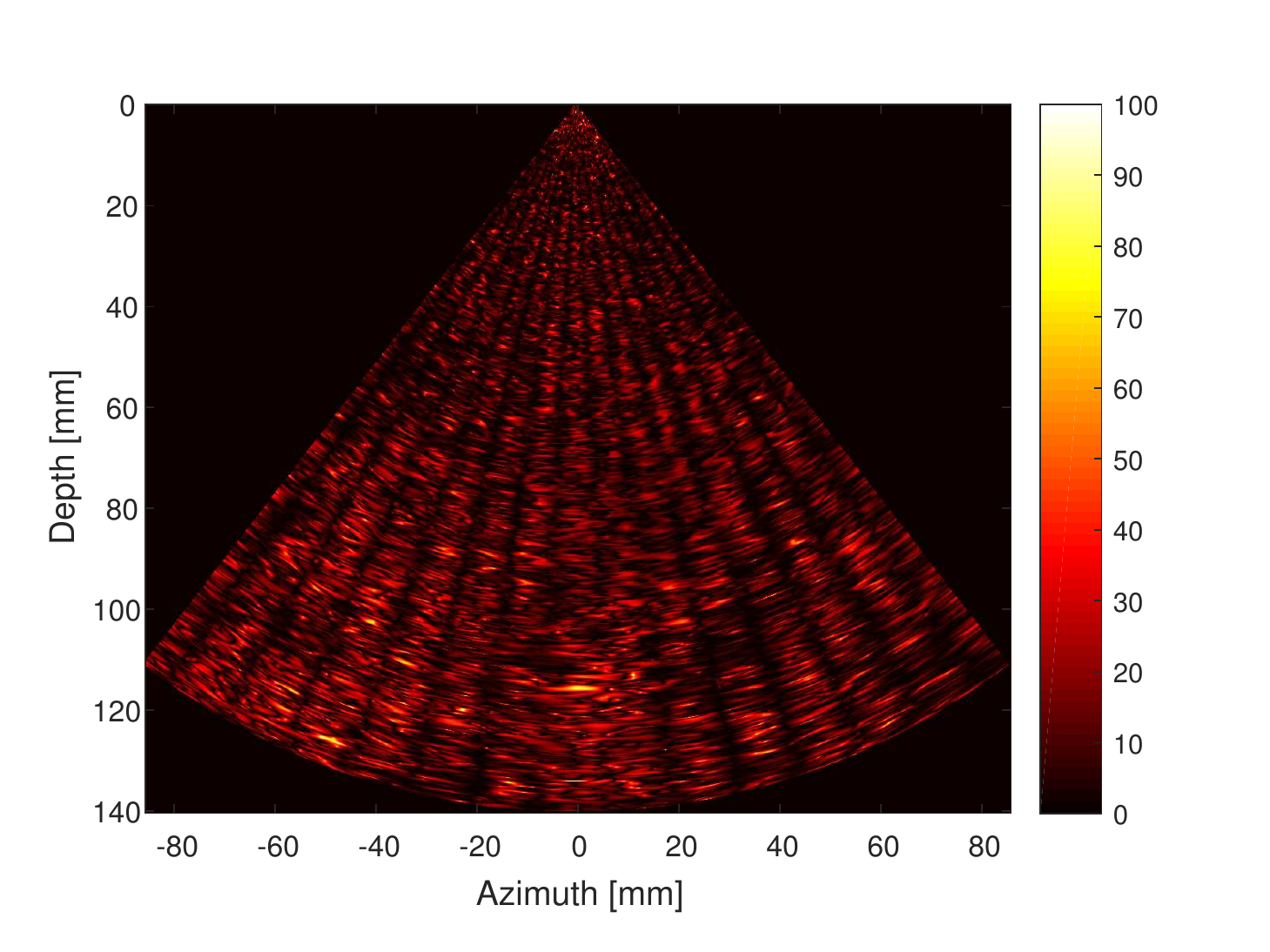} &
		\includegraphics[width = 0.24\textwidth]{results/f2f11_trained20MLA_diff.pdf} 
         \\
              & difference(1.2(b), 1.1(a))  & difference(1.2(c), 1.1(a)) & difference(1.2(d), 1.1(a)) & \\
	\end{tabular}   \\
   \end{minipage}
    \vspace{-0.2cm}
	\caption{\small \textbf{Comparision of Learned Rx vs. Learned Tx-Rx for different initializations.} A test frame from the cineloop comparing the visual results of learned Rx and learned Tx-Rx. The top row depicts the reconstruction obtained from the Learned Rx setting and the third row depicts the reconstruction obtained from the Learned Tx-Rx setting. Even rows depict the difference frames -- difference(x, y) indicates the difference between x and y. The difference maps are scaled between [0-100] for better visualization. Digital zoom-in is recommended.}
	\label{fig:diff_MLA_rates}
\end{sidewaysfigure}

\begin{sidewaysfigure}[!ht]
\begin{minipage}[]{\linewidth}
	\begin{tabular}{ c@{\hskip 0.001\textwidth}c@{\hskip 0.001\textwidth}c@{\hskip 0.001\textwidth}c@{\hskip 0.001\textwidth}c } 
		\includegraphics[width = 0.24\textwidth]{results/f2f11_SLA.pdf} &
		\includegraphics[width = 0.24\textwidth]{results/f2f11_fixed10MLA.pdf} &
		\includegraphics[width = 0.24\textwidth]{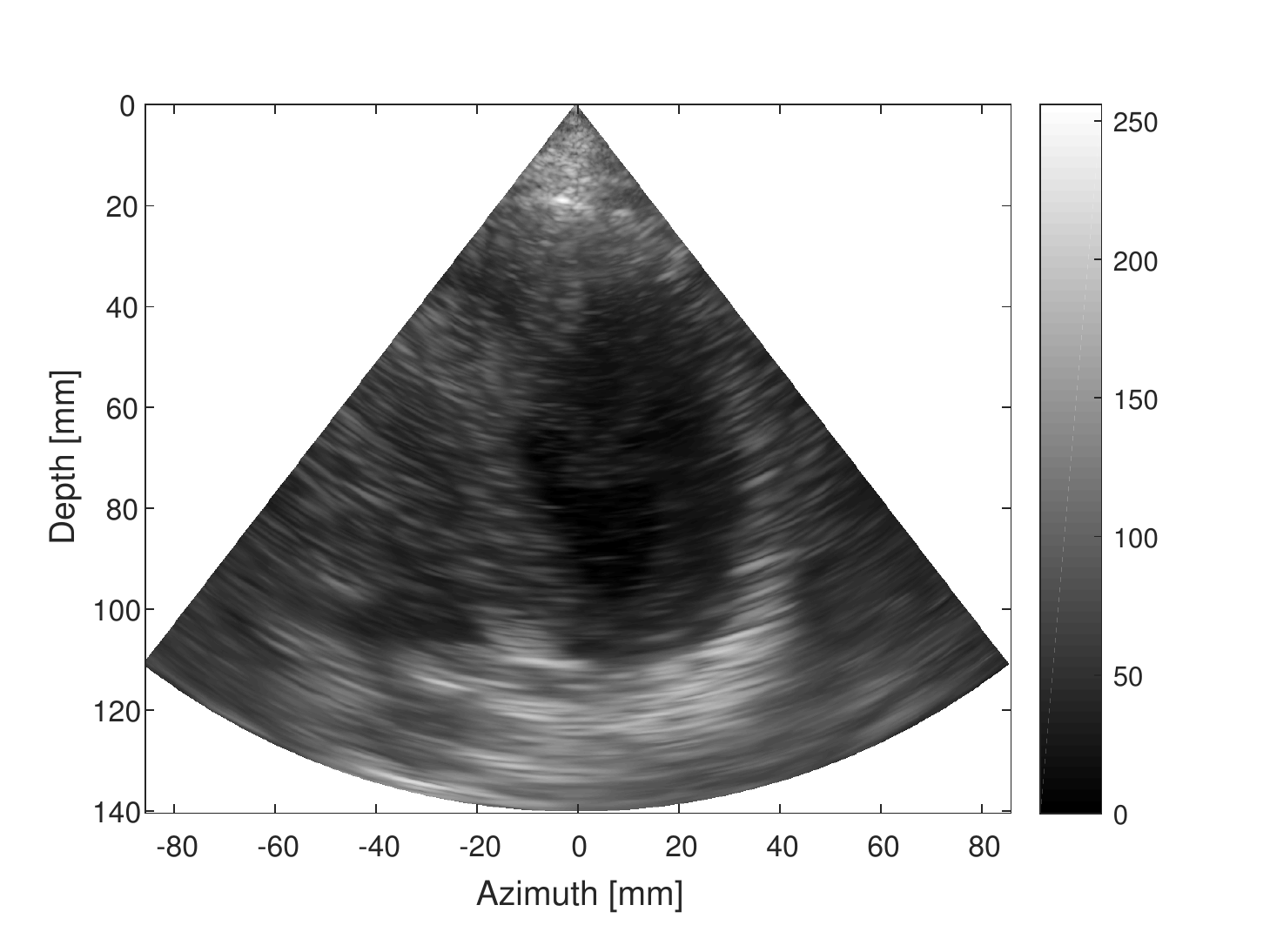} &
	    \includegraphics[width = 0.24\textwidth]{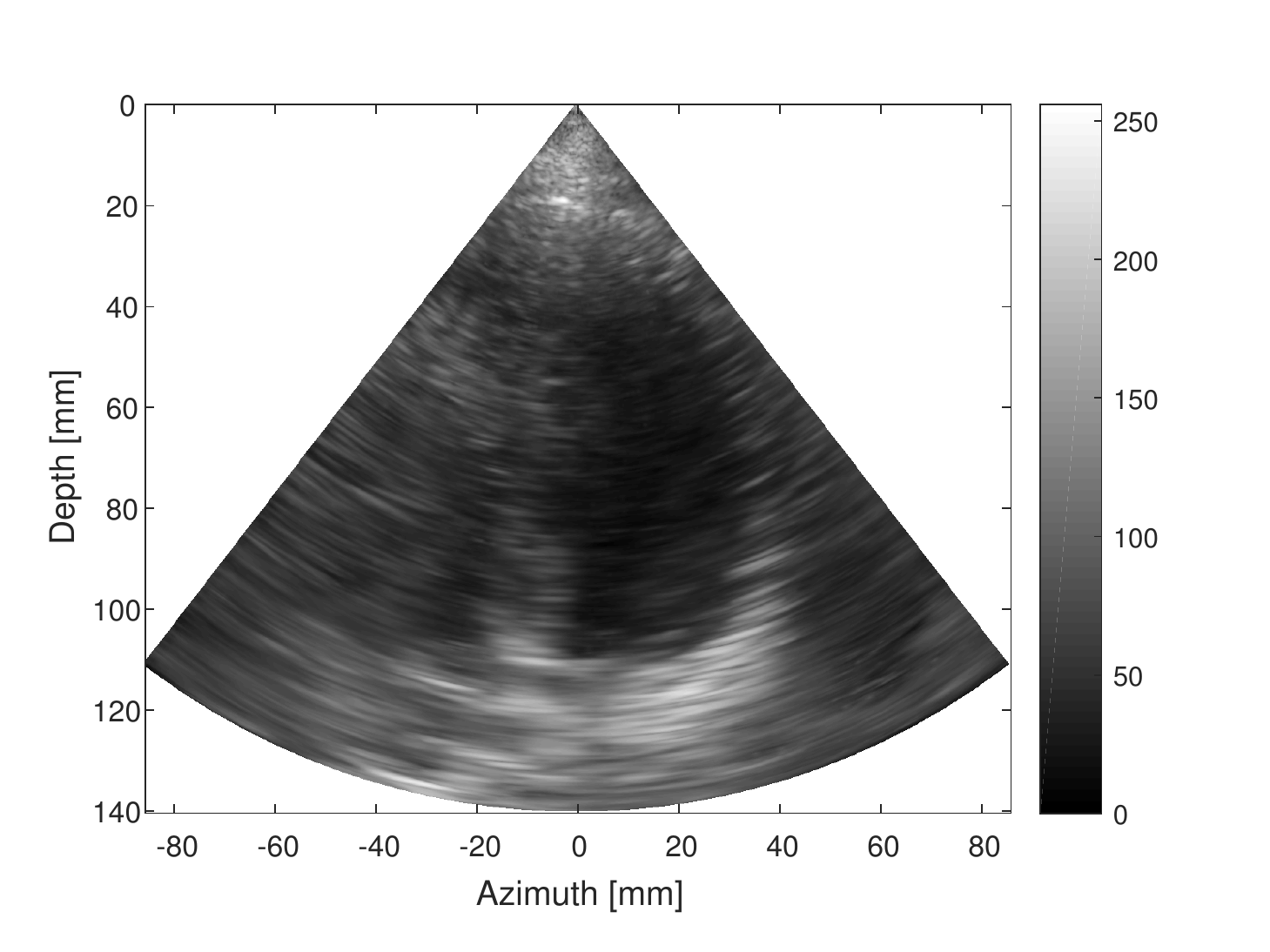} 
	       \\
          2.1(a) SLA   & 2.1(b) Learned Rx 10-MLA  & 2.1(c) Learned Rx 10-MLT & 2.1(d) Learned Rx 10-random & \\
                     &
		\includegraphics[width = 0.24\textwidth]{results/f2f11_fixed10MLA_diff.pdf} &
		\includegraphics[width = 0.24\textwidth]{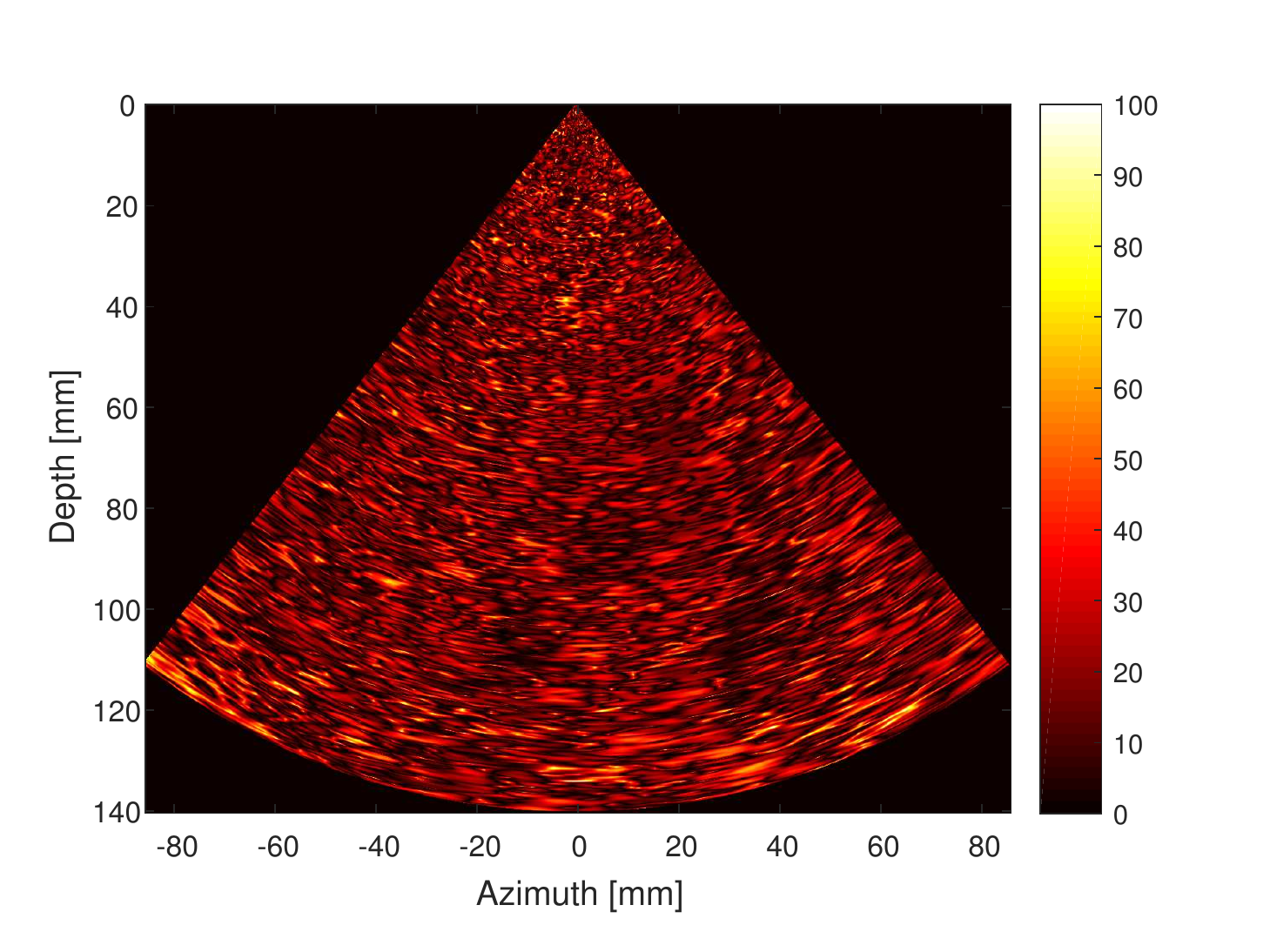} &
		\includegraphics[width = 0.24\textwidth]{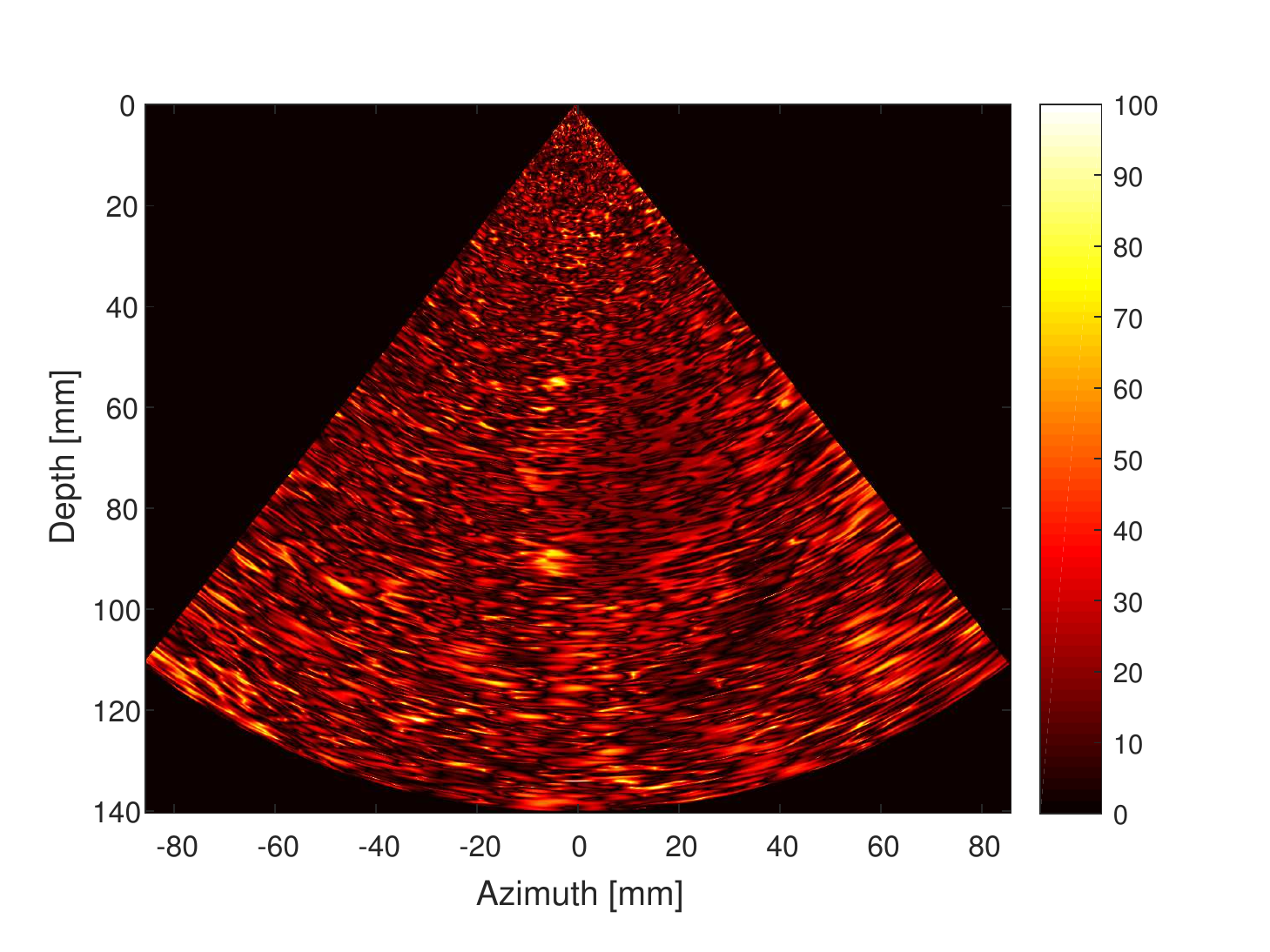} 
         \\
                      & difference(2.1(b), 2.1(a))  & difference(2.1(c), 2.1(a)) &  difference(2.1(d), 2.1(a))& \\
           &
		\includegraphics[width = 0.24\textwidth]{results/f2f11_trained10MLA.pdf} &
		\includegraphics[width = 0.24\textwidth]{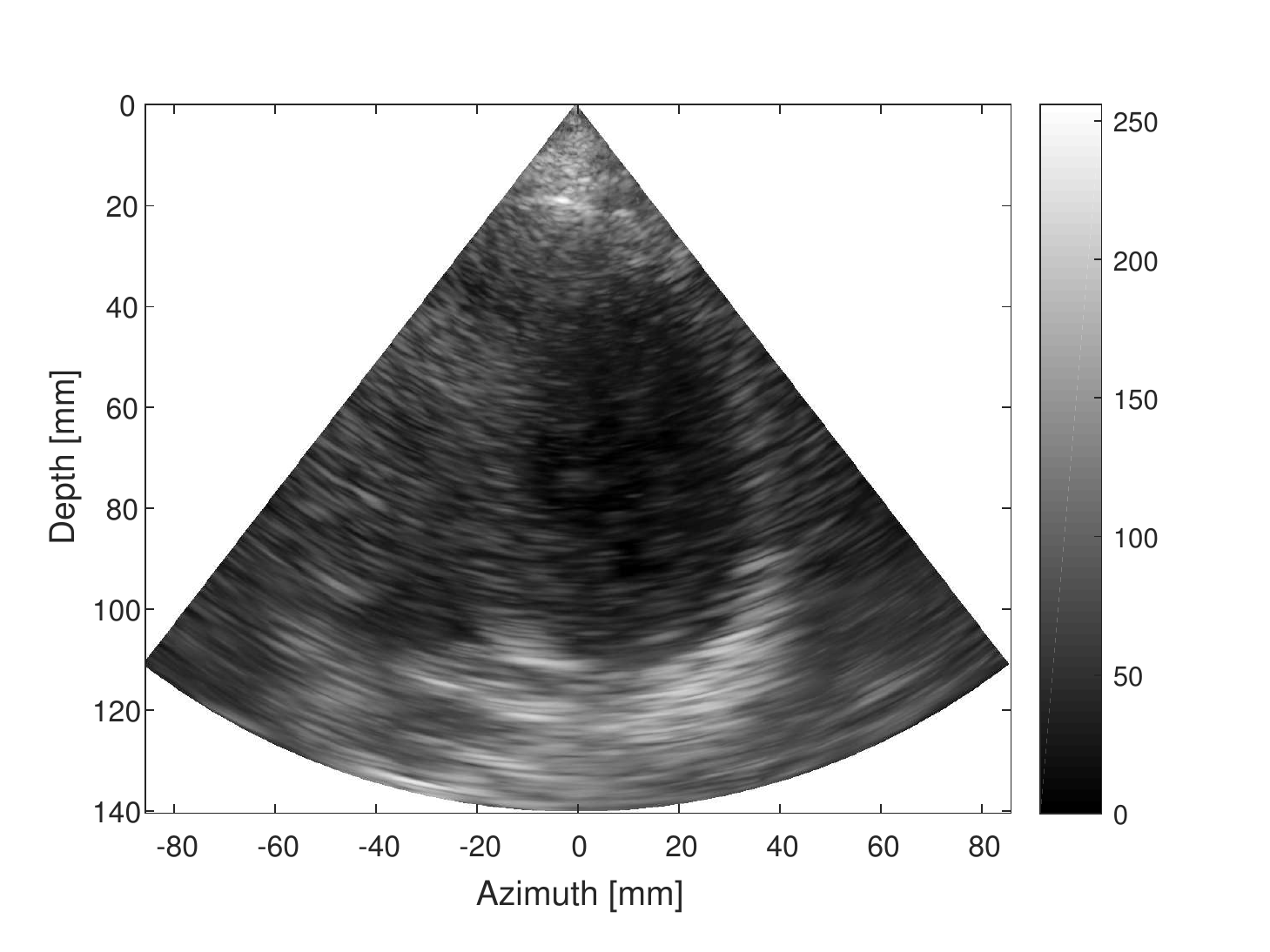} &
		\includegraphics[width = 0.24\textwidth]{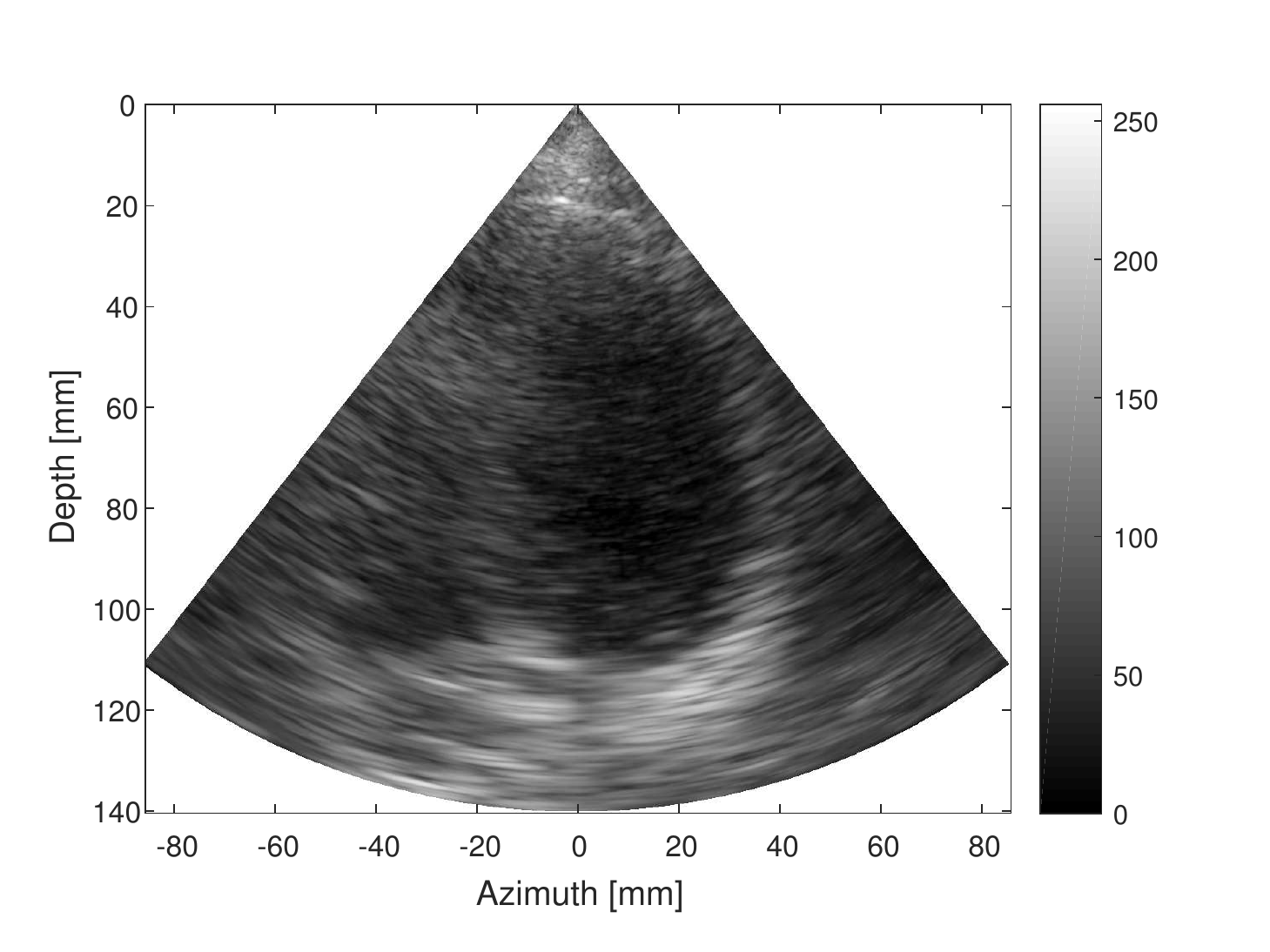} 
         \\
              & 2.2(b) Learned Tx-Rx 10-MLA  & 2.2(c) Learned Tx-Rx 10-MLT & 2.2(d) Learned Tx-Rx 10-random & \\
                         &
		\includegraphics[width = 0.24\textwidth]{results/f2f11_trained10MLA_diff.pdf} &
		\includegraphics[width = 0.24\textwidth]{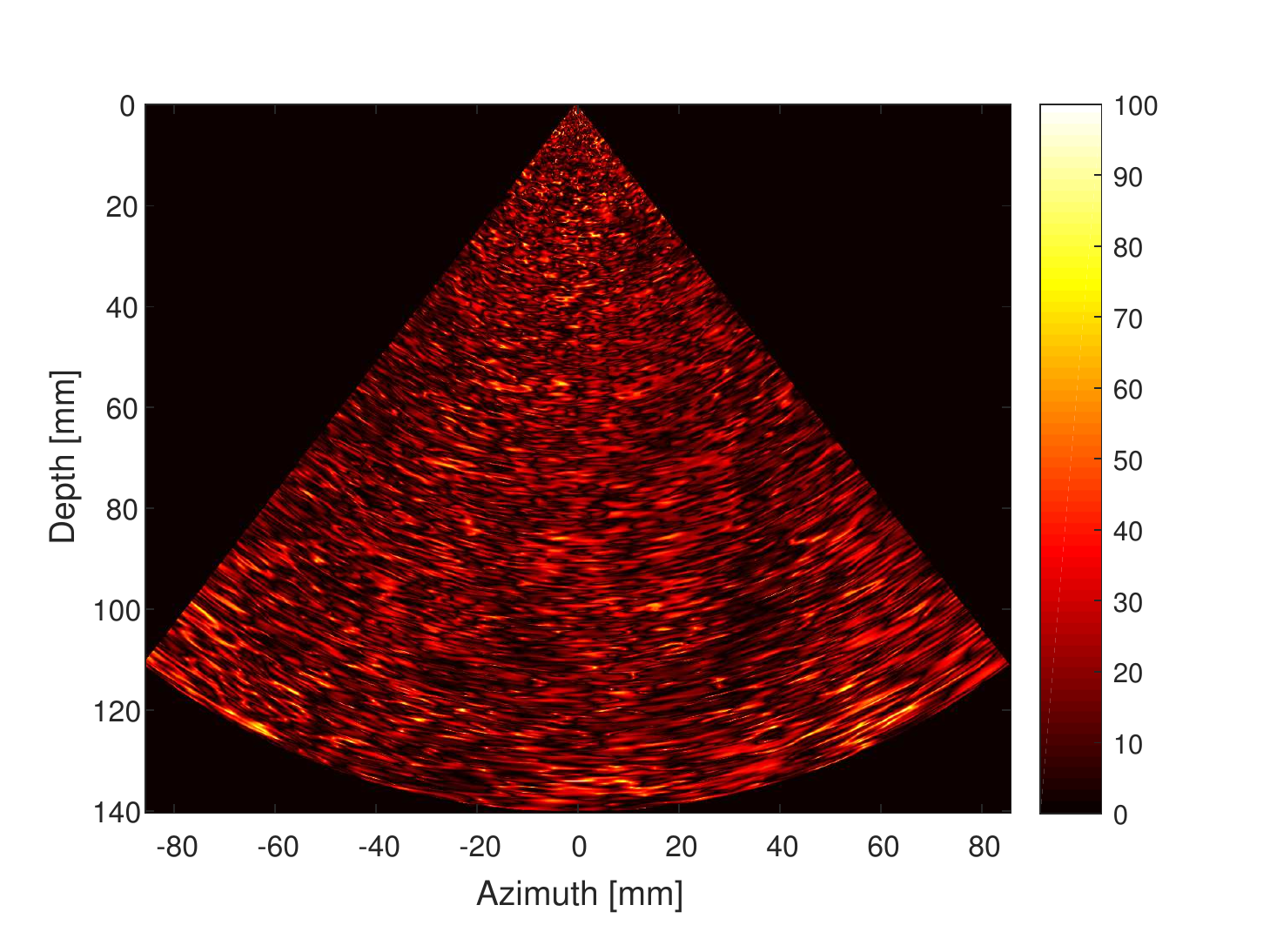} &
		\includegraphics[width = 0.24\textwidth]{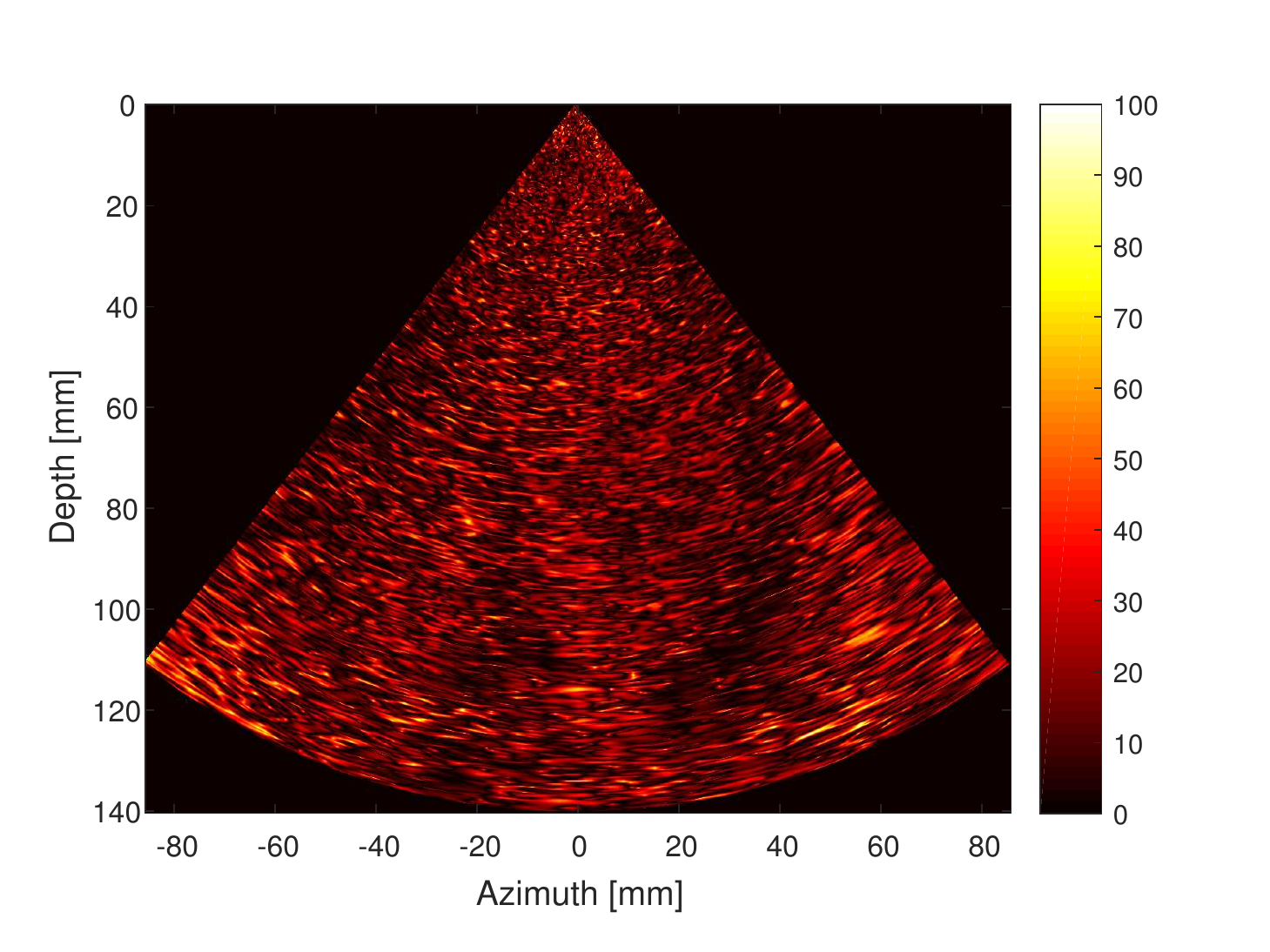} 
         \\
              & difference(2.2(b), 2.1(a))  & difference(2.2(c), 2.1(a)) & difference(2.2(d), 2.1(a)) & \\
    \end{tabular}   \\
   \end{minipage}
    \vspace{-0.2cm}
	\caption{\small \textbf{Comparision of Learned Rx vs. Learned Tx-Rx for different decimation rates.} A test frame from the cineloop comparing the visual results of Learned Rx and Learned Tx-Rx. The top row depicts the reconstruction obtained from the Learned Rx setting and the third row depicts the reconstruction obtained from the Learned Tx-Rx setting. Even rows depict the difference frames -- difference(x, y) indicates the difference between x and y. The difference maps are scaled between [0-100] for better visualization. Digital zoom-in is recommended.}
	\label{fig:diff_inits}
\end{sidewaysfigure}

\begin{table}

\centering
\setlength\tabcolsep{1.5pt} 
\begin{tabular}{l||ll|ll|ll|ll}
        & 7-MLA     &        &       10-MLA &          &       20-MLA &           \\ 
\hline
                & Cr & CNR   & Cr & CNR & Cr & CNR  \\
Learned Rx    & -30.4463dB  & 1.3432   & -33.2432dB   & 1.3453         &-28.3764dB & 1.32  \\
Learned Tx-Rx & -33.2593dB   & 1.3495      & -31.6148dB  & 1.3891        & -32.6599dB  & 1.3214      
\end{tabular}
\caption{Comparison of average contrast-to-noise ratio (CNR) and contrast(Cr) measures between different decimation rates of the transmits. Top and bottom rows indicate the results corresponding to learned Rx and learned Tx-Rx experiment settings respectively.CNR and Cr are calculated for the regions marked within yellow and pink circles drawn in Figure $5, 1.1$(a). }
\label{tab:constrast_tables_diff_inits}
\end{table}

\begin{table}[!ht]

\centering
\setlength\tabcolsep{1.5pt} 
\begin{tabular}{l||ll|ll|ll|ll}
      & 10-MLA     &        &       10-MLT &          &       10-random &           \\ 
\hline
             & Cr & CNR   & Cr & CNR & Cr & CNR  \\
Learned Rx     & -33.2432dB   & 1.3453   & -28.3089 dB   & 1.6155            & -30.3793dB & 1.3452     \\
Learned Tx-Rx  & -31.6148dB  & 1.3891    & -28.8051 dB  &  1.4528            & -31.4859dB & 1.3418       
\end{tabular}

\caption{Comparison of average contrast-to-noise ratio (CNR) and contrast(Cr) measures between different initializations of the transmit patterns. Top and bottom rows indicate the results corresponding to learned Rx and learned Tx-Rx experiment settings respectively. CNR and Cr are calculated for the regions marked within yellow and pink circles drawn in Figure $5, 2.1$(a). }
\label{tab:constrast_tables_diff_rates}
\end{table}

\end{document}